\documentclass[authordate,onecolumn,toc]{miri-tech-article}

\usepackage{amsmath,amssymb,amsthm,amsopn,amsfonts}
\usepackage{mdwlist}
\usepackage[utf8]{inputenc}
\usepackage[T1]{fontenc}
\usepackage{mathtools}
\usepackage{relsize}
\usepackage{lmodern}
\usepackage[ruled,vlined,noend]{algorithm2e}
\usepackage{tikz}
\usepackage[font=small]{caption,subcaption}
\usepackage[colorinlistoftodos,textsize=small,disable]{todonotes}
\usetikzlibrary{shapes,arrows}
\usetikzlibrary{positioning,intersections}
\usepackage{microtype}
\usepackage[e]{esvect}

\DeclareMathOperator*{\argmax}{argmax\;}

\newcommand{\EE}{\ensuremath\mathbb{E}}

\newcommand{\Fig}[1]{figure~\ref{fig:#1}}

\newcommand{\Sec}[1]{section~\ref{sec:#1}}

\newcommand{\Dil}[1]{dilemma~\ref{dil:#1}}

\newcommand{\eq}[1]{\eqref{eq:#1}}

\newcommand{\Eqn}[1]{equation~\eq{#1}}
\newtheorem{dilemma}{Dilemma}

\newcommand{\paper}{paper\xspace}

\newcommand{\Do}{\ensuremath{\mathtt{do}}}
\newcommand{\True}{\ensuremath{\mathtt{true}}}
\newcommand{\cond}{\;\middle|\;}
\newcommand{\V}[1]{\text{\textsc{#1}}}

\newcommand{\then}{\hookrightarrow}
\newcommand{\model}{M}
\newcommand{\Cfor}[1]{\model\textsuperscript{$#1\!\then$}}

\DeclareSymbolFont{symbolsC}{U}{txsyc}{m}{n}
\DeclareMathSymbol{\strictif}{\mathrel}{symbolsC}{74}
\DeclareMathSymbol{\boxright}{\mathrel}{symbolsC}{128}

\newcommand{\Var}[1]{\text{\textsc{#1}}}
\newcommand{\Val}[1]{\text{\textit{#1}}}
\newcommand{\Act}{\Var{Act}\xspace}
\newcommand{\Obs}{\Var{Obs}\xspace}
\newcommand{\Outcome}{\Var{Outcome}\xspace}
\newcommand{\Util}{\ensuremath{\mathcal U}\xspace}
\newcommand{\VV}{\Var{V}\xspace}
\newcommand{\act}{\ensuremath{a}\xspace}
\newcommand{\obs}{\ensuremath{x}\xspace}

\newcommand{\FDTvar}{\Var{fdt}}
\newcommand{\FDTmx}[1][\underline{\obs}]{\ensuremath{\FDTvar(\underline{\model}, #1)}}
\newcommand{\FDTpg}{\ensuremath{\FDTvar(\underline{\Pf}, \underline{\Gf})}}
\newcommand{\FDTpgv}[1]{\ensuremath{\FDTvar(\underline{\Pf}, \underline{\Gf}, \Val{#1})}}
\newcommand{\Pe}{P^\mathit{E}}
\newcommand{\Pc}{P^\mathit{C}}
\newcommand{\Pf}{P^\mathit{F}}
\newcommand{\Gc}{G^\mathit{C}}
\newcommand{\Gf}{G^\mathit{F}}

\newcommand{\EDT}{\mathrm{EDT}}
\newcommand{\CDT}{\mathrm{CDT}}
\newcommand{\FDT}{\mathrm{FDT}}

\SetKwFunction{Universe}{Universe}
\SetKwFunction{PBDT}{PBDT}
\SetKwFunction{sorted}{sorted}
\SetKwProg{Fn}{def}{:}{end}
\SetKwIF{If}{ElseIf}{Else}{if}{:}{elif}{else:}{end}
\SetKwFor{For}{for}{:}{end}
\DontPrintSemicolon

\title{Functional Decision Theory: 
\break
A New Theory of Instrumental Rationality}
\date{}
\author{Eliezer Yudkowsky \and Nate Soares \\ Machine Intelligence Research Institute \\ \{nate,eliezer\}@intelligence.org
}

\begin{document}

\maketitle

\begin{abstract}
This \paper describes and motivates a new decision theory known as \emph{functional decision theory} (FDT), as distinct from causal decision theory and evidential decision theory. Functional decision theorists hold that the normative principle for action is to treat one's decision as the output of a fixed mathematical function that answers the question, ``Which output of this very function would yield the best outcome?'' Adhering to this principle delivers a number of benefits, including the ability to maximize wealth in an array of traditional decision-theoretic and game-theoretic problems where CDT and EDT perform poorly. Using one simple and coherent decision rule, functional decision theorists (for example) achieve more utility than CDT on Newcomb's problem, more utility than EDT on the smoking lesion problem, and more utility than both in Parfit's hitchhiker problem. In this \paper, we define FDT, explore its prescriptions in a number of different decision problems, compare it to CDT and EDT, and give philosophical justifications for FDT as a normative theory of decision-making.

\end{abstract}

\section{Overview} \label{sec:overview}

There is substantial disagreement about which of the two standard decision theories on offer is better: causal decision theory (CDT), or evidential decision theory (EDT). Measuring by utility achieved on average over time, CDT outperforms EDT in some well-known dilemmas \citep{Gibbard:1978}, and EDT outperforms CDT in others \citep{Ahmed:2014}. In principle, a person could outperform both theories by judiciously switching between them, following CDT on some days and EDT on others. This calls into question the suitability of both theories as theories of normative rationality, as well as their usefulness in real-world applications.

We propose an entirely new decision theory, \emph{functional decision theory} (FDT), that maximizes agents' utility more reliably than CDT or EDT. The fundamental idea of FDT is that correct decision-making centers on choosing the output of a \emph{fixed mathematical decision function}, rather than centering on choosing a physical act. A functional decision theorist holds that a rational agent is one who follows a decision procedure that asks ``Which output of this decision procedure causes the best outcome?'', as opposed to ``Which physical act of mine causes the best outcome?'' (the question corresponding to CDT) or ``Which physical act of mine would be the best news to hear I took?'' (the question corresponding to EDT).

As an example, consider an agent faced with a prisoner's dilemma against a perfect psychological twin:

\begin{dilemma}[Psychological Twin Prisoner's Dilemma] \label{dil:twinpd}
  An agent and her twin must both choose to either ``cooperate'' or ``defect.'' If both cooperate, they each receive \$1,000,000.\footnote{Throughout this \paper, we use dollars to represent the subjective utility of outcomes.} If both defect, they each receive \$1,000. If one cooperates and the other defects, the defector gets \$1,001,000 and the cooperator gets nothing. The agent and the twin know that they reason the same way, using the same considerations to come to their conclusions. However, their decisions are causally independent, made in separate rooms without communication. Should the agent cooperate with her twin?
\end{dilemma}

\noindent An agent following the prescriptions of causal decision theory (a ``CDT agent'') would defect, reasoning as follows: ``My action will not affect that of my twin. No matter what action she takes, I win an extra thousand dollars by defecting. Defection dominates cooperation, so I defect.'' She and her twin both reason in this way, and thus they both walk away with \$1,000 \citep{Lewis:1979}.

By contrast, an FDT agent would cooperate, reasoning as follows: ``My twin and I follow the same course of reasoning---this one. The question is how this very course of reasoning should conclude. If it concludes that cooperation is better, then we both cooperate and I get \$1,000,000. If it concludes that defection is better, then we both defect and I get \$1,000. I would be personally better off in the former case, so this course of reasoning hereby concludes \Val{cooperate}.''

On its own, the ability to cooperate in the twin prisoner's dilemma is nothing new. EDT also prescribes cooperation, on the grounds that it would be good news to learn that one had cooperated. However, FDT's methods for achieving cooperation are very unlike EDT's, bearing more resemblance to CDT's methods. For that reason, FDT is able to achieve the benefits of both CDT and EDT, while avoiding EDT's costs (a proposition that we will defend throughout this \paper).

As we will see in \Sec{formalizations}, the prescriptions of FDT agree with those of CDT and EDT in simple settings, where events correlate with the agent's action iff they are caused by the agent's action. They disagree when this condition fails, such as in settings with multiple agents who base their actions off of predictions about each other's decisions. In these cases, FDT outperforms both CDT and EDT, as we will demonstrate by examining a handful of Newcomblike dilemmas.

FDT serves to resolve a number of longstanding questions in the theory of rational choice. It functions equally well in single-agent and multi-agent scenarios, providing a unified account of normative rationality for both decision theory and game theory. FDT agents attain high utility in a host of decision problems that have historically proven challenging to CDT and EDT: FDT outperforms CDT in Newcomb's problem; EDT in the smoking lesion problem; and both in Parfit's hitchhiker problem. They resist extortion in blackmail dilemmas, and they form successful voting coalitions in elections (while denying that voting is irrational). They assign non-negative value to information, and they have no need for ratification procedures or binding precommitments---they can simply adopt the optimal predisposition on the fly, whatever that may be.

Ideas reminiscent of FDT have been explored by many, including \citet{Spohn:2012,Meacham:2010,Yudkowsky:2010:TDT,Dai:2009,Drescher:2006,Gauthier:1994}. FDT has been influenced by \citet{Joyce:1999}, and borrows his representation theorem. It should come as no surprise that an agent can outperform both CDT and EDT as measured by utility achieved; this has been known for some time \citep{Gibbard:1978}. Our contribution is a single simple decision rule that allows an agent to do so in a principled and routinized manner. FDT does not require ad-hoc adjustments for each new decision problem one faces; and with this theory comes a new normative account of counterfactual reasoning that sheds light on a number of longstanding philosophical problems. FDT's combination of theoretical elegance, practical feasibility, and philosophical plausibility makes it a true alternative to CDT and EDT.

In sections~\ref{sec:newcomb}-\ref{sec:parfit}, we informally introduce FDT and compare it to CDT and EDT in three classic Newcomblike problems. In sections~\ref{sec:formalizations} and~\ref{sec:revisit} we provide a more technical account of the differences between these three theories, lending precision to our argument for FDT. In sections~\ref{sec:edt} and~\ref{sec:cdt} we diagnose the failings of EDT and CDT from a new perspective, and show that the failings of CDT are more serious than is generally supposed. Finally, in sections~\ref{sec:perspective} and~\ref{sec:conclusions}, we discuss philosophical motivations and open problems for functional decision theory.

\section{Newcomb's Problem and the Smoking Lesion Problem} \label{sec:newcomb}

Consider the following well-known dilemma, due to \citet{Nozick:1969}:

\begin{dilemma}[Newcomb's Problem] \label{dil:newcomb}
An agent finds herself standing in front of a transparent box labeled ``A'' that contains \$1,000, and an opaque box labeled ``B'' that contains either \$1,000,000 or \$0. A reliable predictor, who has made similar predictions in the past and been correct 99\% of the time, claims to have placed \$1,000,000 in box B iff she predicted that the agent would leave box A behind. The predictor has already made her prediction and left. Box B is now empty or full. Should the agent take both boxes (``two-boxing''), or only box B, leaving the transparent box containing \$1,000 behind (``one-boxing'')?
\end{dilemma}

\noindent The standard formulation of EDT prescribes one-boxing \citep{Gibbard:1978}.\footnote{Specifically, EDT prescribes one-boxing unless the agent assigns prior probability~1 to the hypothesis that she will two-box. See \Sec{revisit} for details.} Evidential decision theorists reason: ``It would be good news to learn that I had left the \$1,000 behind; for then, with 99\% probability, the predictor will have filled the opaque box with \$1,000,000.'' Reasoning thus, EDT agents will be quite predictable, the predictor will fill the box, and EDT agents will reliably walk away \$1,000,000 richer.

CDT prescribes two-boxing. Causal decision theorists reason: ``The predictor already made her prediction at some point in the past, and filled the boxes accordingly. Box B is sitting right in front of me! It already contains \$1,000,000, or is already empty. My decisions now can't change the past; so I should take both boxes and not throw away a free \$1,000.'' Reasoning thus---and having been predicted to reason thus---the CDT agent typically comes away with only \$1,000.

If the story ended there, then we might conclude that EDT is the correct theory of how to maximize one's utility. In Skyrms' smoking lesion problem \citeyearpar{Skyrms:1980}, however, the situation is reversed:

\begin{dilemma}[The Smoking Lesion Problem] \label{dil:lesion}
An agent is debating whether or not to smoke. She knows that smoking is correlated with an invariably fatal variety of lung cancer, but the correlation is (in this imaginary world) entirely due to a common cause: an arterial lesion that causes those afflicted with it to love smoking and also (99\% of the time) causes them to develop lung cancer. There is no direct causal link between smoking and lung cancer. Agents without this lesion contract lung cancer only 1\% of the time, and an agent can neither directly observe nor control whether she suffers from the lesion. The agent gains utility equivalent to \$1,000 by smoking (regardless of whether she dies soon), and gains utility equivalent to \$1,000,000 if she doesn't die of cancer. Should she smoke, or refrain?
\end{dilemma}

\noindent Here, CDT outperforms EDT \citep{Gibbard:1978,Egan:2007}. Recognizing that smoking cannot affect lung cancer, the CDT agent smokes. The EDT agent forgoes the free 1,000 utility and studiously avoids smoking, reasoning: ``If I smoke, then that is good evidence that I have a condition that also causes lung cancer; and I would hate to learn that I have lung cancer far more than I would like to learn that I smoked. So, even though I cannot \emph{change} whether I have cancer, I will select the more auspicious option.''\footnote{The tickle defense of \citet{Eells:1984} or the ratification procedure of \citet{Jeffrey:1983} can be used to give a version of EDT that smokes in this problem (and two-boxes in Newcomb's problem). However, the smoking lesion problem does reveal a fundamental weakness in EDT, and in \Sec{edt} we will examine problems where EDT fails for similar reasons, but where ratification and the tickle defense don't help.}

Imagine an agent that is going to face first Newcomb's problem, and then the smoking lesion problem. Imagine measuring them in terms of utility achieved, by which we mean measuring them by how much utility \emph{we} expect them to attain, on average, if they face the dilemma repeatedly. The sort of agent that we'd expect to do best, measured in terms of utility achieved, is the sort who one-boxes in Newcomb's problem, and smokes in the smoking lesion problem. But \citet{Gibbard:1978} have argued that rational agents \emph{can't} consistently both one-box and smoke: they must either two-box and smoke, or one-box and refrain---options corresponding to CDT and EDT, respectively. In both dilemmas, a background process that is entirely unaffected by the agent's physical action determines whether the agent gets \$1,000,000; and then the agent chooses whether or not to take a sure \$1,000. The smoking lesion problem appears to punish agents that attempt to affect the background process (like EDT), whereas Newcomb's problem appears to punish agents that \emph{don't} attempt to affect the background process. At a glance, this appears to exhaust the available actions. How are we to square this with the intuition that the agent that \emph{actually} does best is the one that one-boxes and smokes?

The standard answer is to argue that, contra this intuition, one-boxing is in fact irrational, and that Newcomb's predictor rewards irrationality \citep{Joyce:1999}. The argument goes, roughly, as follows: Refraining from smoking is clearly irrational, and one-boxing is irrational by analogy---just as an agent can't cause the lesion to appear by smoking, an agent can't cause the amount of money in the box to change after the predictor has left. We shouldn't think less of her for being put in a situation where agents with rational predispositions are punished.

In the context of a debate between two-boxing smokers and one-boxing refrainers, this is a reasonable enough critique of the one-boxing refrainer. Functional decision theorists, however, deny that these exhaust the available options. \citet{Yudkowsky:2010:TDT} provides an initial argument that it is possible to both one-box and smoke using the following thought experiment: Imagine offering a CDT agent two separate binding precommitments. The first binds her to take one box in Newcomb's problem, and goes into effect before the predictor makes her prediction. The second binds her to refuse to smoke in the smoking lesion problem, and goes into effect before the cancer metastasizes. The CDT agent would leap at the first opportunity, and scorn the second \citep{Burgess:2004}. The first precommitment causes her to have a much higher chance of winning \$1,000,000; but the second does not cause her to have a higher chance of survival. Thus, despite the apparent similarities, the two dilemmas must be different in a decision-relevant manner.

Where does the difference lie? It lies, we claim, in the difference between a carcinogenic lesion and a predictor. Newcomb's predictor builds an accurate model of the agent and reasons about her behavior; a carcinogenic lesion does no such thing. If the predictor is reliable, then there is a sense in which the predictor's prediction depends on which action the agent will take in the future; whereas we would not say that the lesion's tendency to cause cancer depends (in the relevant sense) on whether the agent smokes. For a functional decision theorist, this asymmetry makes all the difference.

A functional decision theorist thinks of her decision process as an implementation of a fixed mathematical decision function---a collection of rules and methods for taking a set of beliefs and goals and selecting an action. She weighs her options by evaluating different hypothetical scenarios in which her decision function takes on different \emph{logical outputs}. She asks, not ``What if I used a different process to come to my decisions?'', but: ``What if \emph{this very decision process} produced a different conclusion?''

\citet{Lewis:1979} has shown that the psychological twin prisoner's dilemma is decision-theoretically isomorphic to Newcomb's problem. In that dilemma, the agent's decision process is a function implemented by both prisoners. When a prisoner following FDT imagines the world in which she defects, she imagines a world in which the function she is computing outputs \Val{defect}; and since her twin is computing the same decision function, she assumes the twin will also defect in that world. Similarly, when she visualizes cooperating, she visualizes a world in which her twin cooperates too. Her decision is sensitive to the fact that her twin is a \emph{twin}; this information is not tossed out.

In the same way, the FDT agent in Newcomb's problem is sensitive to the fact that her predictor is a \emph{predictor}. In this case, the predictor is not literally implementing the same decision function as the agent, but the predictor does contain an accurate \emph{mental representation} of the FDT agent's decision function. In effect, the FDT agent has a ``twin'' inside the predictor's mind. When the FDT agent imagines a world where she two-boxes, she visualizes a world where the predictor did not fill the box; and when she imagines a world where she one-boxes, she imagines a world where the box is full. She bases her decision solely on the appeal of these hypothetical worlds, and one-boxes.

By assumption, the predictor's decision (like the twin's decision) reliably corresponds to the agent's decision. The agent and the predictor's model of the agent are like two well-functioning calculators sitting on opposite sides of the globe. The calculators may differ in design, but if they are both well-functioning, then they will output equivalent answers to arithmetical questions.

Just as an FDT agent does not imagine that it is possible for $6288+1048$ to sum to one thing this week and another thing next week, she does not imagine that it is possible for her decision function to, on the same input, have one output today and another tomorrow. Thus, she does not imagine that there can be a difference between her action today and a sufficiently competent prediction about that action yesterday. In Newcomb's problem, when she weighs her options, the only scenarios she considers as possibilities are ``I one-box and the box is 99\% likely to be full'' and ``I two-box and the box is 99\% likely to be empty.'' The first seems more appealing to her, so she one-boxes.

By contrast, she's happy to imagine that there can be a difference between whether or not she smokes, and whether or not the cancer manifests. That correlation is merely statistical---the cancer is not evaluating the agent's future decisions to decide whether or not to manifest. Thus, in the smoking lesion problem, when she weighs her options, the scenarios that she considers as possibilities are ``I smoke (and have probability $p$ of cancer)'' and ``I refrain (and have the same probability $p$ of cancer).'' In this case, smoking seems better, so she smokes.

By treating representations of predictor-like things and lesion-like things differently in the scenarios that she imagines, an FDT agent is able to one-box on Newcomb's problem and smoke in the smoking lesion problem, using one simple rule that has no need for ratification or other probability kinematics.

We can then ask: Given that one is \emph{able} to one-box in Newcomb's problem, without need for precommitments, \emph{should} one?

The standard defense of two-boxing is that Newcomb's problem rewards irrationality. Indeed, it is always possible to construct dilemmas that reward bad decision theories. As an example, we can imagine a decision rule that says to pick the action that comes earliest in the alphabet (under some canonical encoding). In most dilemmas, this rule does poorly; but the rule fares well in scenarios where a reliable predictor rewards exactly the agents that follow this rule, and punishes everyone else. The causal decision theorist can argue that Newcomb's problem is similarly constructed to reward EDT and FDT agents. On this view, we shouldn't draw the general conclusion from this that one-boxing is correct, any more than we should draw the general conclusion from the alphabet dilemma that it is correct to follow ``alphabetical decision theory.''

EDT's irrationality, however, is independently attested by its poor performance in the smoking lesion problem. Alphabetical decision theory is likewise known antecedently to give bad answers in most other dilemmas. ``This decision problem rewards irrationality'' may be an adequate explanation of why an otherwise flawed theory performs well in isolated cases, but it would be question-begging to treat this as a stand-alone argument against a new theory. Here, the general rationality of one-boxing (and of FDT) is the very issue under dispute.

More to the point, the analogy between the alphabet dilemma and Newcomb's problem is tenuous. Newcomb's predictor is not filling boxes according to \emph{how} the agent arrives at a decision; she is only basing her action on a prediction of the decision itself. While it is appropriate to call dilemmas unfair when they directly reward or punish agents for their decision procedures, we deny that there is anything unfair about rewarding or punishing agents for predictions about their \emph{actions}. It is one thing to argue that agents have no say in what decision procedure they implement, and quite another thing to argue that agents have no say in what action they output. In short, Newcomb's problem doesn't punish rational agents; it punishes \emph{two-boxers}. All an agent needs to do to get the high payout is predictably take one box. Thus, functional decision theorists claim that Newcomb's problem is fair game. We will revisit this notion of fairness in \Sec{perspective}.

\section{Subjunctive Dependence} \label{sec:dependence}

The basic intuition behind FDT is that there is some respect in which predictor-like things depend upon an agent's future action, and lesion-like things do not. We'll call this kind of dependence \emph{subjunctive dependence} to distinguish it from (e.g.) straightforward causal dependence and purely statistical dependence.

How does this kind of dependence work in the case of Newcomb's problem? We can assume that the predictor builds a representation of the agent, be it a mental model, a set of notes on scratch-paper, or a simulation \emph{in silico}. She then bases her prediction off of the properties of that representation. In the limiting case where the agent is deterministic and the representation is perfect, the model will always produce the same action as the agent. The behavior of the predictor and agent are interdependent, much like the outputs of two perfectly functioning calculators calculating the same sum.

If the predictor's model is imperfect, or if the agent is nondeterministic, then the interdependence is weakened, but not eliminated. If one observes two calculators computing $6288+1048$, and one of them outputs the number $7336$, one can be fairly sure that the other will also output $7336$. One may not be certain---it's always possible that one hallucinated the number, or that a cosmic ray struck the calculator's circuitry in just the right way to change its output. Yet it will still be the case that one can reasonably infer things about one calculator's behavior based on observing a different calculator. Just as the outputs of the calculator are logically constrained to be equivalent insofar as the calculator made no errors, the prediction and the action in Newcomb's problem are logically constrained to be equivalent insofar as the the predictor made no errors.

When two physical systems are computing the same function, we will say that their behaviors ``subjunctively depend'' upon that function.\footnote{Positing a logical object that decision-makers implement (as opposed to just saying, for example, that interdependent decision-makers depend upon \emph{each other}) is intended only to simplify exposition. We will maintain agnosticism about the metaphysical status of computations, universals, logical objects, etc.}
At first glance, subjunctive dependence may appear to be rooted in causality via the mechanism of a common cause. Any two calculators on Earth are likely to owe their existence to many of the same historical events, if you go back far enough. On the other hand, if one discovered two well-functioning calculators performing the same calculation on different sides of the universe, one might instead suspect that humans and aliens had independently discovered the axioms of arithmetic. We can likewise imagine two different species independently developing (and adopting) the same normative decision theory. In these circumstances, we might speak of universals, laws of nature, or logical or mathematical structures that underlie or explain the convergence; but the relationship is not necessarily ``causal.''

In fact, causal dependence is a special case of subjunctive dependence. Imagine that there is a scribe watching a calculator in Paris, waiting to write down what it outputs. What she writes is causally dependent on the output of the calculator, and also subjunctively dependent upon the output of the calculator---for if it output something different, she would write something different.

Mere statistical correlation, in contrast, does not entail subjunctive dependence. If the Parisian calculator is pink, and (by coincidence) another calculator in Tokyo is also pink, then there is no subjunctive dependency between their colors: if the scribe had bought a green calculator instead, the one in Tokyo would still be pink.

Using this notion of subjunctive dependence, we can define FDT by analogy: \emph{Functional decision theory is to subjunctive dependence as causal decision theory is to causal dependence.}

A CDT agent weighs her options by considering scenarios where her action changes, and events causally downstream from her action change, but everything else is held fixed. An FDT agent weighs her options by considering scenarios where the output of her decision function changes, and events that subjunctively depend on her decision function's output change, but everything else is held fixed. In terms of Peirce's type-token distinction \citeyearpar{Peirce:1934}, we can say that a CDT agent intervenes on the token ``my action is ~\act,'' whereas an FDT agent intervenes on the \emph{type}.

If a certain decision function outputs \Val{cooperate} on a certain input, then it does so of logical necessity; there is no possible world in which it outputs \Val{defect} on that input, any more than there are possible worlds where $6288+1048 \neq 7336$. The above notion of subjunctive dependence therefore requires FDT agents to evaluate counterpossibilities, in the sense of \citet{Cohen:1990}, where the antecedents run counter-to-logic. At first glance this may seem undesirable, given the lack of a satisfactory account of counterpossible reasoning. This lack is the main drawback of FDT relative to CDT at this time; we will discuss it further in \Sec{formalizations}.

In attempting to avoid this dependency on counterpossible conditionals, one might suggest a variant FDT$'$ that asks not ``What if my decision function had a different output?'' but rather ``What if I made my decisions using a different decision function?'' When faced with a decision, an FDT$'$ agent would iterate over functions $f_n$ from some set $\mathcal F$, consider how much utility she would achieve if she implemented that function instead of her actual decision function, and emulate the best $f_n$. Her actual decision function $d$ is the function that iterates over $F$, and $d \not \in \mathcal F$.

However, by considering the behavior of FDT$'$ in Newcomb's problem, we see that it does not save us any trouble. For the predictor predicts the output of $d$, and in order to preserve the desired correspondence between predicted actions and predictions, FDT$'$ cannot simply imagine a world in which she implements $f_n$ instead of $d$; she must imagine a world in which all predictors of $d$ predict as if $d$  behaved like $f_n$---and then we are right back to where we started, with a need for some method of predicting how an algorithm would behave if (counterpossibly) $d$ behaved different from usual.

Instead of despairing at the dependence of FDT on counterpossible reasoning, we note that the difficulty here is technical rather than philosophical. Human mathematicians are able to reason quite comfortably in the face of uncertainty about logical claims such as ``the twin prime conjecture is false,'' despite the fact that either this sentence or its negation is likely a contradiction, demonstrating that the task is not impossible. Furthermore, FDT agents do not need to evaluate counterpossibilities in full generality; they only need to reason about questions like ``How would this predictor's prediction of my action change if the FDT algorithm had a different output?'' This task may be easier. Even if not, we again observe that human reasoners handle this problem fairly well: humans have some ability to notice when they are being predicted, and to think about the implications of their actions on other people's predictions. While we do not yet have a satisfying account of how to perform counterpossible reasoning in practice, the human brain shows that reasonable heuristics exist.

Refer to \citet{Bennett:1974,Lewis:1986,Cohen:1990,Bjerring:2013,Bernstein:2016,Brogaard:2007} for a sample of discussion and research of counterpossible reasoning. Refer to \citet{Garber:1983,Gaifman:2004,Demski:2012a,Hutter:2013,Aaronson:2013:PhilosophersComplexity,Potyka:2015} for a sample of discussion and research into inference in the face of uncertainty about logical facts.

Ultimately, our interest here isn't in the particular method an agent uses to identify and reason about subjunctive dependencies. The important assumption behind our proposed decision theory is that when an FDT agent imagines herself taking two different actions, she imagines corresponding changes in all and only those things that subjunctively depend on the output of her decision-making procedure. When she imagines switching from one-boxing to two-boxing in Newcomb's problem, for example, she imagines the predictor's prediction changing to match. So long as that condition is met, we can for the moment bracket the question of exactly how she achieves this feat of imagination.

\section{Parfit's Hitchhiker} \label{sec:parfit}

FDT's novelty is more obvious in dilemmas where FDT agents outperform CDT \emph{and} EDT agents, rather than just one or the other. Consider Parfit's hitchhiker problem \citeyearpar{Parfit:1986}:
\begin{dilemma}[Parfit's Hitchhiker Problem] \label{dil:parfit}
An agent is dying in the desert. A driver comes along who offers to give the agent a ride into the city, but only if the agent will agree to visit an ATM once they arrive and give the driver \$1,000. The driver will have no way to enforce this after they arrive, but she does have an extraordinary ability to detect lies with 99\% accuracy. Being left to die causes the agent to lose the equivalent of \$1,000,000. In the case where the agent gets to the city, should she proceed to visit the ATM and pay the driver?
\end{dilemma}

\noindent The CDT agent says no. Given that she has safely arrived in the city, she sees nothing further to gain by paying the driver. The EDT agent agrees: on the assumption that she is already in the city, it would be bad news for her to learn that she was out \$1,000. Assuming that the CDT and EDT agents are smart enough to know what they would do upon arriving in the city, this means that neither can honestly claim that they would pay. The driver, detecting the lie, leaves them in the desert to die \citep{Hintze:2014}.

The prescriptions of CDT and EDT here run contrary to many people's intuitions, which say that the most ``rational'' course of action is to pay upon reaching the city. Certainly if these agents had the opportunity to make binding precommitments to pay upon arriving, they would achieve better outcomes.

Consider next the following variant of Newcomb's problem, due to \citet{Drescher:2006}:

\begin{dilemma}[The Transparent Newcomb Problem] \label{dil:transparent}
  Events transpire as they do in Newcomb's problem, except that this time both boxes are transparent---so the agent can see exactly what decision the predictor made before making her own decision. The predictor placed \$1,000,000 in box B iff she predicted that the agent would leave behind box A (which contains \$1,000) upon seeing that both boxes are full. In the case where the agent faces two full boxes, should she leave the \$1,000 behind?
\end{dilemma}

\noindent Here, the most common view is that the rational decision is to two-box. CDT prescribes two-boxing for the same reason it prescribes two-boxing in the standard variant of the problem: whether or not box B is full, taking the extra \$1,000 has better consequences. EDT also prescribes two-boxing here, because \emph{given} that box B is full, an agent does better by taking both boxes.

FDT, on the other hand, prescribes one-boxing, even when the agent knows for sure that box B is full! We will examine how and why FDT behaves this way in more detail in \Sec{revisit}.

Before we write off FDT's decision here as a misstep, however, we should note that one-boxing in the transparent Newcomb problem \emph{is precisely equivalent to paying the driver in Parfit's hitchhiker problem}.

The driver assists the agent (at a value of \$1,000,000) iff she predicts that the agent will pay \$1,000 upon finding herself in the city. The predictor fills box B (at a value of \$1,000,000) iff she predicts that the agent will leave behind \$1,000 upon finding herself facing two full boxes. Why, then, do some philosophers intuit that we should pay the hitchhiker but take both boxes?

It could be that this inconsistency stems from how the two problems are framed. In Newcomb's problem, the money left sitting in the second box is described in vivid sensory terms; whereas in the hitchhiker problem, dying in the desert is the salient image. Or perhaps our intuitions are nudged by the fact that Parfit's hitchhiker problem has \emph{moral} overtones. You're not just outwitting a predictor; you're betraying someone who saved your life.

Whatever the reason, it is hard to put much faith in decision-theoretic intuitions that are so sensitive to framing effects. One-boxing in the transparent Newcomb problem may seem rather strange, but there are a number of considerations that weigh strongly in its favor.

\paragraph{Argument from precommitment:} CDT and EDT agents would both precommit to one-boxing if given advance notice that they were going to face a transparent Newcomb problem. If it is rational to precommit to something, then it should also be rational to predictably \emph{behave as though} one has precommitted. For practical purposes, it is the action itself that matters, and an agent that predictably acts as she would have precommitted to act tends to get rich.

\paragraph{Argument from the value of information:} It is a commonplace in economic theory that a fully rational agent should never expect receiving new information to make her worse off. An EDT agent would pay for the opportunity to blindfold herself so that she can't see whether box B is full, knowing that that information would cause her harm. Functional decision theorists, for their part, do not assign negative expected value to information (as a side effect of always acting as they would have precommitted to act).

\paragraph{Argument from utility:} One-boxing in the transparent Newcomb problem may look strange, but it \emph{works}. Any predictor smart enough to carry out the arguments above can see that CDT and EDT agents two-box, while FDT agents one-box. Followers of CDT and EDT will therefore almost always see an empty box, while followers of FDT will almost always see a full one. Thus, FDT agents achieve more utility in expectation.\par\bigskip

\noindent Expanding on the argument from precommitment, we note that precommitment requires foresight and planning, and can require the expenditure of resources---relying on ad-hoc precommitments to increase one's expected utility is inelegant, expensive, and impractical. Rather than needing to carry out peculiar rituals in order to achieve the highest-utility outcome, FDT agents simply act as they would have ideally precommitted to act.

Another way of articulating this intuition is that we would expect the correct decision theory to endorse its own use. CDT, however, does not: According to CDT, an agent should (given the opportunity) self-modify to stop using CDT in future dilemmas, but continue using it in any ongoing Newcomblike problems \citep{Arntzenius:2008,Soares:2015:toward}. Some causal decision theorists, such as \citet{Burgess:2004} and Joyce (in a 2015 personal conversation), bite this bullet and hold that temporal inconsistency is rational. We disagree with this line of reasoning, preferring to reserve the word ``rational'' for decision procedures that are endorsed by our best theory of normative action. On this view, a decision theory that (like CDT) advises agents to change their decision-making methodology as soon as possible can be lauded for its ability to recognize its own flaws, but is not a strong candidate for the normative theory of rational choice.

Expanding on the argument from the value of information, we view new information as a tool. It is possible to be fed lies or misleading truths; but if one \emph{expects} information to be a lie, then one can simply disregard the information in one's decision-making. It is therefore alarming that EDT agents can \emph{expect} to suffer from learning more about their environments or dispositions, as described by \citet{Skyrms:1982,Arntzenius:2008}.

Expanding on the final argument, proponents of EDT, CDT, and FDT can all agree that it would be great news to hear that a beloved daughter adheres to FDT, because FDT agents get more of what they want out of life. Would it not then be strange if the correct theory of rationality were some \emph{alternative} to the theory that produces the best outcomes, as measured in utility? (Imagine hiding decision theory textbooks from loved ones, lest they be persuaded to adopt the ``correct'' theory and do worse thereby!)

We consider this last argument---the argument from utility---to be the one that gives the precommitment and value-of-information arguments their teeth. If self-binding or self-blinding \emph{were} important for getting more utility in certain scenarios, then we would plausibly endorse those practices. Utility has primacy, and FDT's success on that front is the reason we believe that FDT is a more useful and general theory of rational choice.

The causal decision theorist's traditional response to this line of reasoning has been to appeal to decision-theoretic dominance. An action $a$ is said to dominate an action $b$ if, holding constant the rest of the world's state, switching from~$b$ to~$a$ is sometimes better (and never worse) than sticking with~$b$. \citet{Nozick:1969} originally framed Newcomb's problem as a conflict between the goal of maximizing utility and the goal of avoiding actions that are dominated by other available actions. If CDT and EDT were the only two options available, a case could be made that CDT is preferable on this framing. Both theories variously fail at the goal of utility maximization (CDT in Newcomb's problem, EDT in the smoking lesion problem), so it would seem that we must appeal to some alternative criterion in order to choose between them; and dominance is an intuitive criterion to fall back on.

As we will see in \Sec{revisit}, however, FDT comes with its own dominance principle (analogous to that of CDT), according to which FDT agents tend to achieve higher utility \emph{and} steer clear of dominated actions. More important than FDT's possession of an alternative dominance principle, however, is that in FDT we at last have a general-purpose method for achieving the best real-world outcomes. If we have a strictly superior decision theory on the metric of utility, then we don't need to fall back on the notion of dominance to differentiate between competing theories.

It is for this reason that we endorse one-boxing in the transparent Newcomb problem. When we imagine ourselves facing two full boxes, we feel some of the intuitive force behind the idea that an agent ``could'' break free of the shackles of determinism and two-box upon seeing that both boxes are full. But in 99 cases out of 100, \emph{the kind of agent} that is inclined to conditionally give in to that temptation will actually find herself staring at an empty box B. Parfit's desert is stacked high with the corpses of such agents.

\section{Formalizing EDT, CDT, and FDT} \label{sec:formalizations}

All three of EDT, CDT, and FDT are expected utility theories, meaning that they prescribe maximizing expected utility, which we can define \mkbibparens{drawing from \citet{Gibbard:1978}}\footnote{Early formalizations of decision theory date back to \citet{Ramsey:1931}, \citet{Von-Neumann:1944}, \citet{Savage:1954}, and \citet{Jeffrey:1983}.} as executing an action $a$ that maximizes
\begin{equation}
  \mathcal{EU}(a) \coloneqq \sum_{j=1}^N P(a \then o_j ;\, \obs) \cdot \Util(o_j),
  \label{eq:umax}
\end{equation}
where $o_1, o_2, o_3 \ldots$ are the possible outcomes from some countable set $\mathcal O$; $a$ is an action from some finite set $\mathcal A$; $\obs$ is an observation history from some countable set $\mathcal X$; ${P(a \then o_j ;\, \obs)}$ is the probability that $o_j$ will obtain in the hypothetical scenario where the action $a$ is executed after receiving observations $\obs$; and $\Util$ is a real-valued utility function bounded in such a way that \eq{umax} is always finite.

By the representation theorem of \citet{Joyce:1999}, any agent with a conditional preference ranking and a conditional likelihood function satisfying Joyce's axioms will \emph{want} to maximize \Eqn{umax}, given some set of constraints on the values of $P(- \then -; \, \obs)$, where those constraints are a free parameter---to quote Joyce, ``decision theories should not be seen as offering competing theories of value, but as disagreeing about the epistemic perspective from which actions are to be evaluated.'' EDT, CDT, and FDT can be understood as competing attempts to maximize expected utility by supplying different ``epistemic perspectives'' in the form of differing interpretations of~``$\then$'', the connective for decision hypotheticals.

According to evidential decision theorists, ``$\then$'' should be interpreted as simple Bayesian conditionalization, with ${P(\act \then o_j; \,\obs)}$ standing for ${P(o_j \mid \obs, \act)}$.\footnote{By $P(o_j \mid \obs, \act)$ we mean $P(\Outcome=o_j \mid \Obs=\obs, \Act=\act)$, where \Outcome is an $\mathcal O$-valued random variable representing the outcome; \Obs is a $\mathcal X$-valued random variable representing the observation history; and \Act is an $\mathcal A$-valued random variable representing the agent's action. We write variable names like \Var{This} and values like \Val{this}. We omit variable names when it is possible to do so unambiguously.} Causal and functional decision theorists, on the other hand, insist that conditionals are not counterfactuals. Consider a simple dilemma where a rational agent must choose whether to pick up a \$1 bill or a \$100 bill (but not both). \emph{Conditional} on her picking up the \$1, she probably had a good reason---maybe there is a strange man nearby buying \$1 bills for \$1,000. But if \emph{counterfactually} she took the \$1 instead of the \$100, she would simply be poorer. Causal and functional decision theorists both agree that rational agents should use counterfactual considerations to guide their actions, though they disagree only about which counterfactuals to consider. \citet{Adams:1970}, \citet{Lewis:1981}, and many others have spilled considerable ink discussing this point, so we will not belabor it.

Since Newcomblike problems invoke predictors who build accurate predictions of different agents' reasoning, we will find it helpful to not only ask which theory an agent follows, but \emph{how} she follows that theory's prescriptions.

In general terms, an agent doing her best to follow the prescriptions of one of these three theories should maintain a world-model $\model$---which might be a collection of beliefs and intuitions in her head, a full Bayesian probability distribution updated on her observations thus far, an approximate distribution represented by a neural network, or something else entirely.

When making a decision, for each action $\act$, the agent should modify $\model$ to construct an object $\Cfor\act$ representing what the world would look like if she took that action.\footnote{In the authors' preferred formalization of FDT, agents actually iterate over \emph{policies} (mappings from observations to actions) rather than actions. This makes a difference in certain multi-agent dilemmas, but will not make a difference in this \paper.} We will think of $\Cfor\act$ as a mental image in the agent's head of how the world might look if she chooses $\act$, though one could also think of it as (e.g.) a table giving probabilities to each outcome $o_j$, or an approximate summary of that table. She should then use each $\Cfor\act$ to calculate a value $V_\act$ representing her expected utility if she takes $\act$, and take the action corresponding to the highest value of $V_\act$.

We call $\Cfor\act$ the agent's ``hypothetical for \act.'' On this treatment, hypotheticals are mental or computational objects used as tools for evaluating the expected utility of actions, not (for example) mind-independent possible worlds. Hypotheticals are not themselves beliefs; they are decision-theoretic instruments constructed from beliefs. We think of hypotheticals like we think of notes on scratch paper: transient, useful, and possibly quite representationally thin.

From this perspective, the three decision theories differ only in two ways: how they prescribe representing $\model$, and how they prescribe constructing hypotheticals $\Cfor\act$ from $\model$. For example, according to EDT, $\model$ should be simply be $P(- \mid x)$, a Bayesian probability distribution describing the agent's beliefs about the world updated on observation; and $\Cfor\act$ should be constructed by conditioning $P(- \mid x)$ on $\act$.

Formally, defining a variable $\VV \coloneqq \Util(\Outcome)$ and writing $\EE$ for expectation with respect to $P$, EDT prescribes the action
\begin{equation} \label{eq:edt}
  \EDT(P, \obs) \coloneqq \argmax_{\act \in \mathcal A} \EE\left(\VV \mid \Obs=\obs, \Act=\act\right),
\end{equation}
where it is understood that if an action \act has probability $0$ then it is not considered. If the agent makes no observations during a decision problem, then we will omit $\obs$ and write, e.g., $\EDT(P)$.

Equation~\eq{edt} can be read in at least three interesting ways. It can be seen as a Joyce-style constraint on the value of $P(\act \then o_j;\,\obs)$. It can be seen as advice saying what information the agent needs to make their world-model $\model$ (a probability distribution) and how they should construct hypotheticals $\Cfor\act$ (Bayesian conditionalization). Finally, it can be seen as a step-by-step procedure or algorithm that could in principle be scrupulously followed by a human actor, or programmed into a decision-making machine: take $P$ and $\obs$ as input, compute ${\EE(\VV\mid\obs, \act)}$ for each action $\act$, and execute the $\act$ corresponding to the highest value.

As an aside, note that \Eqn{edt} does not address the question of how a consistent initial belief state~$P$ is constructed. This may be nontrivial if the agent has beliefs about her own actions, in which case $P$ needs to assign probabilities to claims about $\EDT(P, \obs)$, possibly requiring ratification procedures \`{a} la \citet{Jeffrey:1983}.

According to CDT, a rational agent's $\model$ should be not just a distribution $P$, but $P$ augmented with additional data describing the causal structure of the world; and $\Cfor\act$ should be constructed from $\model$ by performing a causal intervention on $P$ that sets ${\Act=\act}$. There is some debate about exactly how causal structure should be represented and how causal interventions should be carried out, going back to at least \citet{Lewis:1973}. Pearl's graphical approach \citeyearpar{Pearl:1996} is perhaps the most complete account, and it is surely among the easiest to formalize, so we will use it here.\footnote{In this \paper, we use graphical formulations of CDT and FDT because they are simple, formal, and easy to visualize. However, they are not the only way to formalize the two theories, and there are causal and functional decision theorists who don't fully endorse the equations developed in this section. The only features of graphical models that we rely upon are the independence relations that they encode. For example, our argument that CDT two-boxes relies on graphs only insofar as the graph says that the agent's action is causally independent from the prediction of the predictor. Any formalization of CDT (graphical or otherwise) that respects the independence relationships in our graphs will agree with our conclusions.}

In Pearl's formulation, $\model$ is a ``causal theory,'' which, roughly speaking, is a pair $(P, G)$ where $G$ is a graph describing the direction of causation in the correlations between the variables of $P$. To go from $\model$ to $\Cfor\act$, Pearl defines an operator $\Do$. Again speaking roughly, $P(- \mid \Do(\Var{Var}=\Val{val}))$ is a modified version of $P$ in which all variables that are causally downstream from \Var{Var} (according to $G$) are updated to reflect $\Var{Var}=\Val{val}$, and all other variables are left untouched.

\begin{figure}
    \centering
    \begin{tikzpicture}[scale=0.8,every node/.style={transform shape}]
      \node[rounded rectangle,draw, minimum size=1cm] (nState) at (3, 7.5) {\Var{Predisposition}};
      \node[rectangle,draw,minimum size=1cm] (nAct) at (1, 3.5) {\Act};
      \node[rounded rectangle,draw,minimum size=1cm] (nPrediction) at (5, 5) {\Var{Prediction}};
      \node[rounded rectangle,draw,minimum size=1cm] (nNoise) at (6.5, 6.5) {\Var{Accurate}};
      \node[rounded rectangle,draw,minimum size=1cm] (nBoxB) at (5, 3.5) {\Var{Box B}};
      \node[rounded rectangle,double,draw,minimum size=1cm] (nObs) at (3, 3.5) {\Obs};
      \node[rounded rectangle,draw,minimum size=1cm] (nOut) at (3, 2) {\Outcome};
      \node[diamond,draw,minimum size=1cm] (nUtil) at (3, 0.5) {\VV};
      \draw[->, >=latex] (nState) to[out=205,in=90] (nAct);
      \draw[->, >=latex] (nState) to[out=-25,in=90] (nPrediction);
      \draw[->, >=latex] (nNoise) -> (nPrediction);
      \draw[->, >=latex] (nPrediction) -> (nBoxB);
      \draw[->, >=latex] (nAct) -> (nOut);
      \draw[->, >=latex] (nBoxB) -> (nOut);
      \draw[->, >=latex] (nOut) -> (nUtil);
    \end{tikzpicture}
    \caption{A causal graph for CDT agents facing Newcomb's problem. The agent observes the double-bordered node (which is, in this case, unused), intervenes on the rectangular node, and calculates utility using the diamond node.\label{fig:newcomb.cdt}}
\end{figure}
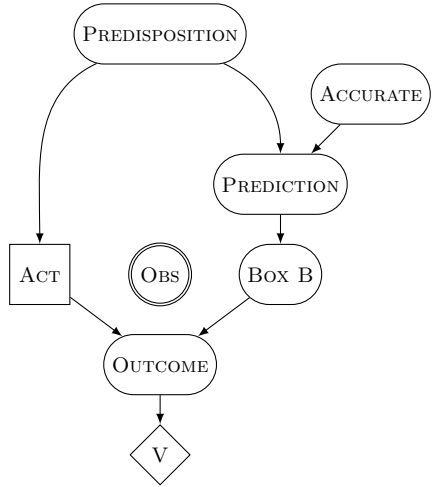

For example, consider \Fig{newcomb.cdt}, which gives a causal graph for Newcomb's problem. It must be associated with a probability distribution $P$ that has a variable for each node in the graph, such that \Act is a function of \Var{Predisposition}, \Outcome is a function of \Act and \Var{Box B}, and so on.

Graphically, the operation $\Do(\Act=\Val{onebox})$ begins by setting $\Act=\Val{onebox}$ without changing any other variables. It then follows arrows outwards, and recomputes the values of any node it finds, substituting in the values of variables that it has affected so far. For example, when it gets to \Outcome, it updates it to $o=\Outcome(\Val{onebox}, \Var{Box B})$; and when it gets to \VV it updates it to $\VV(o)$. Any variable that is not downstream from \Act is unaffected. We can visualize $\Do(\Act=\Val{onebox})$ as severing the correlation between \Act and \Var{Predisposition} and then performing a normal Bayesian update; or we can visualize it as updating \Act and everything that it causally affects while holding everything else fixed. Refer to \citet{Pearl:1996} for the formal details.

Pearl's formulation yields the following equation for CDT:
\begin{equation} \label{eq:cdt}
  \CDT(P, G, \obs) \coloneqq \argmax_{\act \in \mathcal A} \EE\left(\VV \cond \Do(\Act=\act), \Obs=\obs\right).
\end{equation}
As before, this equation can be interpreted as a Joyce-style constraint, or as advice about how to construct hypotheticals from a world-model, or as a step-by-step decision procedure. We leave aside the question of how to construct $G$ from observation and experiment; it is examined in detail by \citet{Pearl:2000}.

In CDT hypotheticals, some correlations in $P$ are preserved but others are broken. For any variable $Z$, correlations between \Act and $Z$ are preserved iff $G$ says that the correlation is caused by \Act---all other correlations are severed by the $\Do$ operator. FDT's hypotheticals are also constructed via surgery on a world-model that preserves some correlations and breaks others. What distinguishes these two theories is that where a CDT agent performs surgery on a variable \Act representing a \emph{physical} fact (about how her body behaves), an FDT agent performs the surgery on a variable representing a \emph{logical} fact (about how her decision function behaves).

For a functional decision theorist, $P$ must contain variables representing the outputs of different mathematical functions. There will for example be a variable representing the value of $6288+1048$, one that (hopefully) has most of its probability on $7336$. A functional decision theorist with world-model $\model$ and observations $\obs$ calculates $\FDT(\model, \obs)$ by intervening on a variable $\FDTmx$ in $\model$ that represents the output of $\FDT$ when run on inputs $\model$ and $\obs$. Here we use underlines to represent dequoting, i.e., if $x \coloneqq 3$ then $\Var{Z}\underline{x}$ denotes the variable name $\Var{Z}3$. Note the self-reference here: the model $\model$ contains variables for many different mathematical functions, and the FDT algorithm singles out a variable $\FDTmx$ in the model whose name depends on the model. This self-reference is harmless: an FDT agent does not need to know the \emph{value} of $\FDT(\model, \obs)$ in order to have a variable $\FDTmx$ \emph{representing} that value, and Kleene's second recursion theorem shows how to construct data structures that contain and manipulate accurate representations of themselves \citep{Kleene:1938}, via a technique known as ``quining''.

Instead of a $\Do$ operator, FDT needs a $\True$ operator, which takes a logical sentence $\phi$ and updates $P$ to represent the scenario where $\phi$ is true. For example, ${P(\Var{Riemann} \mid \True(\lnot\Var{TPC}))}$ might denote the agent's subjective probability that the Riemann hypothesis would be true if (counterfactually) the twin prime conjecture were false. Then we could say that
\[
  \FDT^*(P, \obs)=\argmax_{\act \in \mathcal A} \EE\left(\VV\mid\True(\FDTvar(\underline{P}, \underline{x})=\act)\right).
\]
Unfortunately, it's not clear how to define a $\True$ operator. Fortunately, we don't have to. Just as CDT requires that $P$ come augmented with information about the causal structure of the world, FDT can require that $P$ come augmented with information about the logical, mathematical, computational, causal, etc. structure of the world more broadly. Given a graph $G$ that tells us how changing a logical variable affects all other variables, we can re-use Pearl's $\Do$ operator to give a decision procedure for FDT:\footnote{This is not the only way to formalize a $\True$ operator. Some functional decision theorists hope that the study of counterpossibilities will give rise to a method for conditioning a distribution on logical facts, allowing one to define $\FDT(P, \obs) = \argmax_{\!\!\act} \EE\left(\VV \cond \FDTvar(\underline{M}, \underline{\obs})=\act\right),$ an evidential version of FDT. We currently lack a formal definition of conditional probabilities that can be used with false logical sentences (such as $\FDTvar(\underline{M}, \underline{\obs})=a_2$ when in fact it equals $a_1$). Thus, for the purposes of this paper, we require that the relevant logical dependency structure be given as an input to FDT, in the same way that the relevant causal structure is given as an input to CDT.}
\begin{equation} \label{eq:fdt}
  \FDT(P, G, \obs) \coloneqq \argmax_{\act \in \mathcal A} \EE\left(\VV \cond \Do\left(\FDTvar(\underline P,\underline G,\underline x)= \act\right)\right).
\end{equation}

Comparing equations~\eq{cdt} and~\eq{fdt}, we see that there are two differences between FDT and CDT. First, where CDT intervenes on a node $\Act$ representing the physical action of the agent, FDT intervenes on a node $\FDTvar(\underline P, \underline G, \underline x)$ representing the outputs of its decision procedure given its inputs. Second, where CDT responds to observation by Bayesian conditionalization, FDT responds to observation by changing which node it intervenes upon. When CDT's observation history updates from $x$ to $y$, CDT changes from conditioning its model on $\Obs=x$ to conditioning its model on $\Obs=y$, whereas FDT changes from intervening on the variable $\FDTvar(\underline P, \underline{G}, \underline x)$ to $\FDTvar(\underline P, \underline{G}, \underline y)$ instead. We will examine the consequences of these two differences in the following section.

Equation~\eq{fdt} is sufficient for present purposes, though it rests on shakier philosophical foundations than \eq{cdt}. \citet{Pearl:2000} has given a compelling philosophical account of how to deduce the structure of causation from observation and experiment, but no such formal treatment has yet been given to the problem of deducing the structure of other kinds of subjunctive dependence. Equation~\eq{fdt} works \emph{given} a graph that accurately describes how changing the value of a logical variable affects other variables, but it is not yet clear how to construct such a thing---nor even whether it can be done in a satisfactory manner within Pearl's framework. Figuring out how to deduce the structure of subjunctive dependencies from observation and experiment is perhaps the largest open problem in the study of functional decision theory.\footnote{An in-depth discussion of this issue is beyond the scope of this paper, but refer to \Sec{dependence} for relevant resources.}

In short, CDT and FDT both construct counterfactuals by performing a surgery on their world-model that breaks some correlations and preserves others, but where CDT agents preserve only causal structure in their hypotheticals, FDT agents preserve all decision-relevant subjunctive dependencies in theirs. This analogy helps illustrate that Joyce's representation theorem applies to FDT as well as CDT. Joyce's representation theorem \citeyearpar{Joyce:1999} is very broad, and applies to any decision theory that prescribes maximizing expected utility relative to a set of constraints on an agent's beliefs about what would obtain under different conditions. To quote \citet{Joyce:1999}:
\begin{quote}
  It should now be clear that all expected utility theorists can agree about the broad foundational assumptions that underlie their common doctrine. [\,\ldots] Since the constraints on conditional preferences and beliefs needed to establish the existence of conditional utility representations in Theorem 7.4 are common to both the causal and evidential theories, there is really no difference between them as far as their core accounts of valuing are concerned. [\,\ldots] There remains, of course, an important difference between the causal and evidential approaches to decision theory. Even though they agree about the way in which prospects should be valued once an epistemic perspective is in place, the two theories differ about the correct epistemic perspective from which an agent should evaluate his or her potential actions.
\end{quote}
He is speaking mainly of the relationship between CDT and EDT, but the content applies just as readily to the relationship between FDT and CDT. FDT is defined, by \eq{fdt}, as an expected utility theory which differs from CDT only in what constraints it places on an agent's thoughts about what would obtain if (counterfactually) she made different observations and/or took different actions. In particular, where CDT requires that an agent's hypotheticals respect causal constraints, FDT requires also that the agent's counterfactuals respect logical constraints. FDT, then, is a cousin of CDT that endorses the theory of expected utility maximization and meets the constraints of Joyce's representation theorem. The prescriptions of FDT differ from those of CDT only on the dimension that Joyce left as a free parameter: the constraints on how agents think about the hypothetical outcomes of their actions.

From these equations, we can see that all three theories agree on models $(P, G)$ in which all correlations between $\Act$ and other variables are caused (according to $G$) by $\Act$, except perhaps $\FDTvar(\underline P, \underline G)$, on which \Act may subjunctively depend (according to $G$). In such cases,
\[
  \EE[V \mid \act] = \EE[V \mid \Do(\act)] = \EE[V \mid \Do(\FDTvar(\underline P, \underline G)=\act)],
\]
so all three equations produce the same output. However, this condition is violated in cases where events correlate with the agent's action in a manner that is not caused by the action, which happens when, e.g., some other actor is making predictions about the agent's behavior. It is for this reason that we turn to Newcomblike problems to distinguish between the three theories, and demonstrate FDT's superiority, when measuring in terms of utility achieved.

\section{Comparing the Three Decision Algorithms' Behavior} \label{sec:revisit}

With equations for EDT, CDT, and FDT in hand, we can put our analyses of Newcomb's problem, the smoking lesion problem, and the transparent Newcomb problem on a more formal footing. We can construct probability distributions and graphical models for a given dilemma, feed them into our equations, and examine precisely what actions an agent following a certain decision algorithm would take, and why.

In what follows, we will consider the behavior of three agents---Eve, Carl, and Fiona---who meticulously follow the prescriptions of equations~\eq{edt},~\eq{cdt}, and~\eq{fdt} respectively. We will do this by defining $P$ and $G$ objects, and evaluating the $\EDT$, $\CDT$, and $\FDT$ algorithms on those inputs. Note that our $P$s and $G$s will describe an agent's \emph{models} of what situation they are facing, as opposed to representing the situations themselves. When Carl changes a variable $Z$ in his distribution $\Pc{}$, he is not affecting the object that $Z$ represents in the world; he is manipulating a representation of $Z$ in his head to help him decide what to do. Also, note that we will be evaluating agents in situations where their models accurately portray the situations that they face. It is possible to give a Eve agent a model $\Pe{}$ that says she's playing checkers when in fact she's playing chess, but this wouldn't tell us much about her decision-making skill.

We will evaluate Eve, Carl, and Fiona using the simplest possible world-models that accurately capture a given dilemma. In the real world, their models would be much more complicated, containing variables for each and every one of their beliefs. In this case, solving equations~\eq{edt},~\eq{cdt}, and~\eq{fdt} would be intractable: maximizing over all possible actions is not realistic. We study the idealized setting because we expect that bounded versions of Eve, Carl, and Fiona would exhibit similar strengths and weaknesses relative to one another. After all, an agent approximating EDT will behave differently than an agent approximating CDT, even if both agents are bounded and imperfect.

\subsection{Newcomb's Problem}

To determine how Eve the EDT agent responds to Newcomb's problem, we need a distribution $\Pe{}$ describing the epistemic state of Eve when she believes she is facing Newcomb's problem. We will use the distribution in \Fig{newcomb.net}, which is complete except for the value $\Pe{}(\Val{onebox})$, Eve's prior probability that she one-boxes.

\begin{figure}
    \centering
    \begin{subfigure}{\textwidth}
      \centering
      \begin{tikzpicture}[scale=1,every node/.style={transform shape}]
        \node[rounded rectangle,draw, minimum size=1cm] (nState)   at (3, 7.5) {\Var{Predisposition}};
        \node[rounded rectangle,draw,minimum size=1cm] (nAct)   at (1, 3.5) {\Act};
        \node[rounded rectangle,draw,minimum size=1cm] (nPrediction) at (5, 5) {\Var{Prediction}};
        \node[rounded rectangle,draw,minimum size=1cm] (nNoise) at (6.5, 6.5) {\Var{Accurate}};
        \node[rounded rectangle,draw,minimum size=1cm] (nBoxB) at (5, 3.5) {\Var{Box B}};
        \node[rounded rectangle,draw,minimum size=1cm] (nOut)   at (3, 2) {\Outcome};
        \node[diamond,draw,minimum size=1cm] (nUtil)   at (3, 0.5) {\VV};
        \draw[->, >=latex] (nAct) to[out=90,in=205] (nState);
        \draw[->, >=latex] (nState) to[out=-25,in=90] (nPrediction);
        \draw[->, >=latex] (nNoise) -> (nPrediction);
        \draw[->, >=latex] (nPrediction) -> (nBoxB);
        \draw[->, >=latex] (nAct) -> (nOut);
        \draw[->, >=latex] (nBoxB) -> (nOut);
        \draw[->, >=latex] (nOut) -> (nUtil);
      \end{tikzpicture}
      \par\bigskip
      \bgroup
      \def\arraystretch{1.5}
      \begin{tabular}{r|c|}
        \multicolumn{1}{c}{} & \multicolumn{1}{c}{\Act}\\
        \cline{2-2}
        \Val{onebox} & ? \\
        \cline{2-2}
        \Val{twobox} & ? \\
        \cline{2-2}
      \end{tabular}
      \hfill
      \begin{tabular}{r|c|}
        \multicolumn{1}{c}{} & \multicolumn{1}{c}{\Var{Accurate}}\\
        \cline{2-2}
        \Val{accurate} & 0.99 \\
        \cline{2-2}
        \Val{inaccurate} & 0.01 \\
        \cline{2-2}
      \end{tabular}
      \egroup
      \begin{align*}
        \Var{Predisposition}(\Val{onebox}) & \coloneqq \Val{oneboxer} &
          \hfill &
          \Var{Outcome}(\Val{twobox}, \Val{full}) \coloneqq \Val{2f} \\
        \Var{Predisposition}(\Val{twobox}) & \coloneqq \Val{twoboxer} &
          \hfill &
          \Var{Outcome}(\Val{onebox}, \Val{full}) \coloneqq \Val{1f} \\
        \Var{Prediction}(\Val{oneboxer}, \Val{accurate}) & \coloneqq \Val{1} &
          \hfill &
          \Var{Outcome}(\Val{twobox}, \Val{empty}) \coloneqq \Val{2e} \\
        \Var{Prediction}(\Val{twoboxer}, \Val{accurate}) & \coloneqq \Val{2} &
          \hfill &
          \Var{Outcome}(\Val{onebox}, \Val{empty}) \coloneqq \Val{1e} \\
        \Var{Prediction}(\Val{oneboxer}, \Val{inaccurate}) & \coloneqq \Val{2} &
          \hfill &
          \VV(\Val{2f}) \coloneqq \Val{\$1,001,000} \\
        \Var{Prediction}(\Val{twoboxer}, \Val{inaccurate}) & \coloneqq \Val{1} &
          \hfill &
          \VV(\Val{1f}) \coloneqq \Val{\$1,000,000} \\
          \Var{Box B}(\Val{1}) & \coloneqq \Val{full} &
          \hfill &
          \VV(\Val{2e}) \coloneqq \Val{\$1,000} \\
          \Var{Box B}(\Val{2}) & \coloneqq \Val{empty} &
          \hfill &
          \VV(\Val{1e}) \coloneqq \Val{\$0} \\
      \end{align*}
    \end{subfigure}
    \caption{A Bayesian network for Newcomb's problem. The only stochastic nodes are \Act and \Var{Accurate}; all the other nodes are deterministic. As you may verify, the results below continue to hold if we add more stochasticity, e.g., by making \Var{Predisposition} strongly but imperfectly correlated with \Act. The prior probabilities on \Act are left as a free parameter. This graph is more verbose than necessary: we could collapse \VV and \Var{Outcome} into a single node, and collapse \Var{Prediction} and \Var{Box B} into a single node. Note that this is \emph{not} a causal graph; the arrow from \Act to \Var{Predisposition} describes a correlation in the agent's beliefs but does not represent causation.\label{fig:newcomb.net}}
\end{figure}

Eve's behavior depends entirely upon this value. If she were certain that she had a one-boxing predisposition, then $\Pe{}(\Val{onebox})$ would be $1$. She would then be unable to condition on $\Act=\Val{twobox}$, because two-boxing would be an event of probability zero. As a result, she would one-box. Similarly, if she were certain she had a two-boxing predisposition, she would two-box. In both cases, her prior would be quite accurate---it would assign probability~1 to her taking the action that she would in fact take, given that prior. As noted by \citet{Spohn:1977}, given extreme priors, Eve can be made to do \emph{anything}, regardless of what she knows about the world.\footnote{The ratification procedure of \citet{Jeffrey:1983} and the meta-tickle defense of \citet{Eells:1984} can in fact be seen as methods for constructing \smash{$\Pe{}$} that push \smash{$\Pe{}(\Val{onebox})$} to~0, causing Eve to two-box.} Eve's choices are at the mercy of her priors in a way that Carl and Fiona's are not---a point to which we'll return in \Sec{edt}.

But assume her priors are not extreme, i.e., that ${0 < \Pe{}(\Val{onebox}) < 1}.$ In this case, Eve solves \Eqn{edt}, which requires calculating the expectation of \VV in two different hypotheticals, one for each available action. To evaluate one-boxing, she constructs a hypothetical by conditioning $\Pe{}$ on \Val{onebox}. The expectation of \VV in this hypothetical is \$999,000, because ${\Pe{}(\Val{full}\mid\Val{onebox}) = 0.99}$.
To evaluate two-boxing, she constructs a second hypothetical by conditioning $\Pe{}$ on \Val{twobox}. In this case, expected utility is \$11,000, because ${\Pe{}(\Val{full}\mid\Val{twobox}) = 0.01}$.
The value associated with one-boxing is higher, so $\EDT(\Pe)=\Val{onebox}$, so Eve one-boxes.

In words, Eve reasons: ``Conditional on one-boxing, I very likely have a one-boxing predisposition; and one-boxers tend to get rich; so I'd probably be a gets-rich sort of person. That would be great! Conditional on two-boxing, though, I very likely have a two-boxing disposition; two-boxers tend to become poor; so I'd probably be a stays-poor sort of person. That would be worse, so I one-box.'' The predictor, seeing that Eve assigns nonzero prior probability to one-boxing and following this very chain of reasoning, can easily see that Eve one-boxes, and will fill the box. As a result, Eve will walk away rich.

What about Carl? Carl's probability distribution $\Pc{}$ also follows \Fig{newcomb.net} (though for him, the variables represent different things---Carl's variable \Var{Predictor} represents a predictor that was thinking about Carl all day, whereas Eve's corresponding variable represents a predictor that was thinking about Eve, and so on.) To figure out how Carl behaves, we need to augment $\Pc{}$ with a graph $\Gc{}$ describing the causal relationships between the variables. This graph is given in \Fig{newcomb.cdt2}. Our task is to evaluate $\CDT(\Pc, \Gc)$. We haven't specified the value $\Pc(\Val{onebox})$, but as we will see, the result is independent from this value.

Like Eve, Carl makes his choice by comparing two hypothetical scenarios. Carl constructs his first hypothetical by performing the causal intervention ${\Do(\Val{onebox})}$. This sets $\Act$ to $\Val{onebox}$, then propagates the update to only the variables causally downstream from \Act. \V{Box B} is causally independent from \Act according to the graph, so the probability of \Val{full} remains unchanged. Write $p$ for this probability. According to Carl's first hypothetical, if he one-boxes then there is a $p$ chance he wins \$1,000,000 and a $(1-p)$ chance he walks away empty-handed. To make his second hypothetical, Carl performs the causal surgery ${\Do(\Val{twobox})}$, which also does not alter the probability of \Val{full}. According to that hypothetical, if Carl two-boxes, he has a $p$ chance of \$1,001,000 and a $(1-p)$ chance of \$1,000.

No matter what the value of $p$ is, \VV is higher by 1000 in the second hypothetical, so $\CDT(\Pc, \Gc)=\Val{twobox}$. Thus, Carl two-boxes. In words, Carl reasons: ``Changing my action does not affect the probability that box B is full. Regardless of whether it's full or empty, I do better by taking box A, which contains a free \$1,000.''

This means that $\Pc{}(\Val{onebox})$ should be close to zero, because any agent smart enough to follow the reasoning above (including Carl) can see that Carl will take two boxes. Furthermore, the predictor will have no trouble following the reasoning that we just followed, and will not fill the box; so Carl will walk away poor.

\begin{figure}
    \centering
    \begin{subfigure}{\textwidth}
      \centering
      \subcaptionbox{Carl's graph $\Gc$ for Newcomb's problem. This graph is merely a simplified version of \Fig{newcomb.cdt}.\label{fig:newcomb.cdt2}}[.45\linewidth]{
      \begin{tikzpicture}[scale=0.8,every node/.style={transform shape}]
        \node[rounded rectangle,draw, minimum size=1cm] (nState) at (3, 7.5) {\Var{Predisposition}};
        \node[rectangle,draw,minimum size=1cm] (nAct) at (1, 3.5) {\Act};
        \node[rounded rectangle,draw,minimum size=1cm] (nPrediction) at (5, 5) {\Var{Prediction}};
        \node[rounded rectangle,draw,minimum size=1cm] (nNoise) at (6.5, 6.5) {\Var{Accurate}};
        \node[rounded rectangle,draw,minimum size=1cm] (nBoxB) at (5, 3.5) {\Var{Box B}};
        \node[diamond,draw,minimum size=1cm] (nUtil) at (3, 2.5) {\VV};
        \draw[->, >=latex] (nState) to[out=205,in=90] (nAct);
        \draw[->, >=latex] (nState) to[out=-25,in=90] (nPrediction);
        \draw[->, >=latex] (nNoise) -> (nPrediction);
        \draw[->, >=latex] (nPrediction) -> (nBoxB);
        \draw[->, >=latex] (nAct) -> (nUtil);
        \draw[->, >=latex] (nBoxB) -> (nUtil);
      \end{tikzpicture}}%
      \hfill
      \subcaptionbox{Fiona's graph $\Gf{}$ for Newcomb's problem. Fiona intervenes on a variable \FDTpg{} representing what the FDT algorithm outputs given Fiona's world-model.\label{fig:newcomb.fdt}}[.45\linewidth]{
      \begin{tikzpicture}[scale=0.8, every node/.style={transform shape}]
        \node[rectangle,draw, minimum size=1cm] (nState)   at (3, 7.5) {\FDTpg};
        \node[rounded rectangle,draw,minimum size=1cm] (nAct) at (1, 3.5) {\Act};
        \node[rounded rectangle,draw,minimum size=1cm] (nPrediction) at (5, 5) {\Var{Prediction}};
        \node[rounded rectangle,draw,minimum size=1cm] (nNoise) at (6.5, 6.5) {\Var{Accurate}};
        \node[rounded rectangle,draw,minimum size=1cm] (nBoxB) at (5, 3.5) {\Var{Box B}};
        \node[diamond,draw,minimum size=1cm] (nUtil) at (3, 2.5) {\VV};
        \draw[->, >=latex] (nState) to[out=205,in=90] (nAct);
        \draw[->, >=latex] (nState) to[out=-25,in=90] (nPrediction);
        \draw[->, >=latex] (nNoise) -> (nPrediction);
        \draw[->, >=latex] (nPrediction) -> (nBoxB);
        \draw[->, >=latex] (nAct) -> (nUtil);
        \draw[->, >=latex] (nBoxB) -> (nUtil);
      \end{tikzpicture}}%
  \end{subfigure}
\end{figure}

Third, we turn our attention to Fiona. Her distribution $\Pf{}$ is similar to those of Eve and Carl, except that instead of reasoning about her ``predisposition'' as a common cause of her act and the predictor's prediction, she reasons about the \emph{decision function} $\FDTpg{}$ that she is implementing. Fiona's graph $\Gf$ (given in \Fig{newcomb.fdt}) is similar to $\Gc$, but she intervenes on the variable \FDTpg{} instead of \Act.

Fiona, like Eve and Carl, weighs her options by comparing two hypotheticals. In her hypotheticals, the value of \Var{Prediction} varies with the value of \Act---because they \emph{both} vary according to the value of \FDTpg{}. To make her first hypothetical, she performs the intervention ${\Do(\FDTpg{}=\Val{onebox})}$, which sets the probability of $\Act=\Val{onebox}$ to~$1$ and the probability of $\Var{Box B}=\Val{full}$ to $0.99$. To make her second hypothetical, she performs ${\Do(\FDTpg{}=\Val{twobox})}$, which sets the probability of $\Act=\Val{onebox}$ to~$0$ and the probability of $\Var{Box B}=\Val{full}$ to $0.01$. Expected utility in the first case is \$990,000; expected utility in the second case is \$11,000. Thus, $\FDT(\Pf,\Gf)=\Val{onebox}$ and Fiona one-boxes.\footnote{Be careful to distinguish \FDTpg{} from $\FDT(\Pf,\Gf)$. The former is a variable in Fiona's model that represents the output of her decision process, which she manipulates to produce an action. The latter is the action produced.}

In English, this corresponds to the following reasoning: ``If this very decision procedure outputs \Val{onebox}, then my body almost surely takes one box and the predictor likely filled box B. If instead this very decision procedure outputs \Val{twobox}, then my body almost surely takes two boxes and the predictor likely left box B empty. Between those two possibilities, I prefer the first, so this decision procedure hereby outputs \Val{onebox}.''

Assuming that Fiona is smart enough to follow the above line of reasoning, $\Pf{}(\FDTpg{}=\Val{onebox})\approx 1$, because FDT agents obviously one-box. Similarly, a predictor capable of following this argument will have no trouble predicting that Fiona always one-boxes---and so Fiona walks away rich.

Here we pause to address a common objection: If Fiona is almost certain that she has a one-boxing disposition (and 99\% certain that box B is full), then upon reflection, won't she decide to take two boxes? The answer is no, because of the way that Fiona weighs her options. To consider the consequences of changing her action, she imagines a hypothetical scenario in which her decision function has a different output. Even if she is quite sure that the box is full because $\FDT(\Pf, \Gf)=\Val{onebox}$, when you ask her what \emph{would happen if} she two-boxed, she says that, for her to two-box, the FDT algorithm would have to output \Val{twobox} on input $(\Pf,\Gf)$. If the FDT algorithm itself behaved differently, then other things about the universe would be different---much like we should expect elliptical curves to have different properties if (counterpossibly) Fermat's last theorem were false as opposed to true. Fiona's graph $\Gf$ tells her how to imagine this counterpossibility, and in particular, because her algorithm and the predictor's prediction subjunctively depend on the same function, she imagines a hypothetical world where most things are the same but box B is probably empty. That imagined hypothetical seems worse to her, so she leaves the \$1,000 behind.

Nowhere in Fiona's reasoning above is there any appeal to a belief in retrocausal physics. If she understands modern physics, she'll be able to tell you that information cannot travel backwards in time. She does not think that a physical signal passes between her action and the predictor's prediction; she just thinks it is foolish to imagine her action changing without also imagining $\FDT(\Pf, \Gf)$ taking on a different value, since she thinks the predictor is good at reasoning about FDT. When she imagines two-boxing, she therefore imagines a hypothetical world where box B is empty.

\subsection{An Aside on Dominance}

Functional decision theorists deny the argument for two-boxing from dominance. The causal decision theorist argues that one-boxing is irrational because, though it tends to make one richer in practice, switching from one-boxing to two-boxing (while holding constant everything except the action's effects) always yields more wealth. In other words, whenever you one-box then it will be revealed that you left behind a free \$1,000; but whenever you two-box you left behind nothing; so (the argument goes) one-boxing is irrational.

But this practice of checking what an agent should have done \emph{holding constant everything except the action's effects} ignores important aspects of the world's structure: when the causal decision theorist asks us to imagine the agent's action switching from \Val{twobox} to \Val{onebox} holding fixed the predictor's prediction, they are asking us to imagine the agent's physical action changing while holding fixed the behavior of the agent's decision function. This is akin to handing us a well-functioning calculator calculating $6288+1048$ and asking us to imagine it outputting $3159$, holding constant the fact that $6288+1048=7336$.

Two-boxing ``dominates'' if dominance is defined in terms of CDT counterfactuals; where regret is measured by visualizing a world where the action was changed but the decision function was not. But this is not an independent argument for CDT; it is merely a restatement of CDT's method for assessing an agent's options.

An analogous notion of ``dominance'' can be constructed using FDT-style counterfactuals, in which action~$a$ dominates action~$b$ if, holding constant all relevant subjunctive dependencies, switching the output of the agent's algorithm from~$b$ to~$a$ is sometimes better (and never worse) than sticking with~$b$. According to this notion of dominance, FDT agents never take a dominated action. In Newcomb's problem, if we hold constant the relative subjunctive dependency (that the predictor's prediction is almost always equal to the agent's action) then switching from one box to two makes the agent worse off.

In fact, \emph{every} method for constructing hypotheticals gives rise to its own notion of dominance. If we define ``dominance'' in terms of Bayesian conditionalization---$a$ ``dominates''~$b$ if ${\EE(\VV\mid a) > \EE(\VV\mid b)}$---then refusing to smoke ``dominates'' in the smoking lesion problem. To assert that one action ``dominates'' another, one must assume a particular method of evaluating counterfactual actions. Every expected utility theory comes with its own notion of dominance, and dominance doesn't afford us a neutral criterion for deciding between candidate theories. For this reason, we much prefer to evaluate decision theories based on how much utility they tend to achieve in practice.

\subsection{The Smoking Lesion Problem}
To model Eve's behavior in the smoking lesion problem, we will need a new distribution describing her beliefs when she faces the smoking lesion problem. The insight of \citet{Gibbard:1978} is that, with a bit of renaming, we can re-use the distribution $\Pe{}$ from Newcomb's problem. Simply carry out the following renaming:
\begin{align*}
  \Var{Predisposition} & \text{ becomes } & \Var{Lesion} \\
  \Val{oneboxer} & \text{ becomes } & \Val{nolesion} \\
  \Val{twoboxer} & \text{ becomes } & \Val{lesion} \\
  \Var{Accurate} & \text{ becomes } & \Var{Luck} \\
  \Val{accurate} & \text{ becomes } & \Val{unlucky} \\
  \Val{inaccurate} & \text{ becomes } & \Val{lucky} \\
  \Var{Prediction} & \text{ becomes } & \Var{Cancer} \\
  \Val{1} & \text{ becomes } & \Val{nocancer} \\
  \Val{2} & \text{ becomes } & \Val{cancer} \\
  \Var{Box B} & \text{ becomes } & \Var{Death} \\
  \Val{empty} & \text{ becomes } & \Val{dead} \\
  \Val{full} & \text{ becomes } & \Val{alive} \\
  \Act & \text{ becomes } & \Act \\
  \Val{twobox} & \text{ becomes } & \Val{smoke} \\
  \Val{onebox} & \text{ becomes } & \Val{refrain}
\end{align*}
and we're good to go. Clearly, then, Eve refrains from smoking: $\EDT(\Pe)=\Val{refrain}$ for the same reason that it was $\Val{onebox}$ in Newcomb's problem. She reasons that most smokers die, and most non-smokers don't, and she'd rather hear that she was in the latter category, so she refrains from smoking.

Similarly, with mere renaming, we can re-use Carl's model for Newcomb's problem to make his model $(\Pc, \Gc)$ for the smoking lesion problem. Carl smokes: $\CDT(\Pc, \Gc)=\Val{smoke}$ for the same reason that it was \Val{twobox} in Newcomb's problem. He reasons that the probability of cancer doesn't depend on his action, so regardless of whether he has the lesion, he's better off smoking.

In Fiona's case, though, her graph in Newcomb's problem cannot be re-used in the smoking lesion problem---the relations of subjunctive dependence differ. According to Fiona's graph for Newcomb's problem, changing \FDTpg{} changes both \Act and \Var{Prediction}, because both of those variables represent something in the world that depends on the output of $\FDT(\Pf,\Gf)$. In her graph for the smoking lesion problem, though, the corresponding variable should be connected to \Act but not \Var{Cancer}, because whether or not the cancer metastasizes does not depend upon the output of the FDT procedure. Carl can re-use his causal graph from Newcomb's problem because he does not track these subjunctive dependencies, but Fiona's behavior depends on these subjunctive dependencies, so her graphs must differ.

In fact, building $\Pf$ and $\Gf$ for the smoking lesion problem requires that we formalize the problem a bit further. According to the problem description, the lesion determines whether or not the agent likes smoking. We can formalize this by saying that there are two utility functions, $\Util_S$ and $\Util_R$, where $\Util_R$ differs from $\Util_S$ in that it values smoking at -\$1. Thus, there are two different distributions Fiona could have: $P_S$, in which $\VV$ is calculated using $\Util_S$; and $P_R$, in which \VV is calculated using $\Util_R$. Fiona can't tell which distribution she actually uses; it's a black box to her.

\begin{figure}
    \centering
    \begin{tikzpicture}[scale=0.8, every node/.style={transform shape}]
      \node[rounded rectangle,draw, minimum size=1cm] (nState)   at (3, 7.5) {\Var{Lesion}};
      \node[rectangle,draw, minimum size=1cm]         (nf1)   at (3, 6) {$\FDTvar(\underline{\Pf_S}, \underline{\Gf})$};
      \node[rounded rectangle,draw, minimum size=1cm] (nf2)   at (3, 4) {$\FDTvar(\underline{\Pf_R}, \underline{\Gf})$};
      \node[rounded rectangle,draw,minimum size=1cm] (nNoise) at (6.5, 6.5) {\Var{Luck}};
      \node[rounded rectangle,draw,minimum size=1cm] (nAct)   at (1, 3.5) {\Act};
      \node[rounded rectangle,draw,minimum size=1cm] (nPrediction) at (5, 5) {\Var{Cancer}};
      \node[rounded rectangle,draw,minimum size=1cm] (nBoxB) at (5, 3.5) {\Var{Death}};
      \node[diamond,draw,minimum size=1cm] (nUtil)   at (3, 2.5) {\VV};
      \draw[->, >=latex] (nState) to[out=210,in=90] (nAct);
      \draw[->, >=latex] (nState) to[out=-30,in=90] (nPrediction);
      \draw[->, >=latex] (nPrediction) -> (nBoxB);
      \draw[->, >=latex] (nf1) -> (nAct);
      \draw[->, >=latex] (nf2) -> (nAct);
      \draw[->, >=latex] (nAct) -> (nUtil);
      \draw[->, >=latex] (nBoxB) -> (nUtil);
      \draw[->, >=latex] (nNoise) -> (nPrediction);
    \end{tikzpicture}
    \caption{Fiona's graph $\Gf$ for the smoking lesion problem. In the case where she has the lesion, the point of intervention will be $\FDTvar(\underline{\Pf_S}, \underline{\Gf})$, as illustrated. In the case where she does not have the lesion, the point of intervention will instead be $\FDTvar(\underline{\Pf_R}, \underline{\Gf})$. In either case, Fiona does not have introspective access to the point of intervention, and computes her action under ignorance about her own preferences.\label{fig:lesion.fdt}}
\end{figure}

Because there are two world-models she might be running, there are also two decision functions she could be running: $\FDT(\Pf_S, \Gf)$ or $\FDT(\Pf_R, \Gf)$. Which one she runs depends upon whether she has the lesion. Thus, in her graph, the node $\Act$ depends on $\Var{Lesion}$, $\FDTvar(\underline{\Pf_S}, \underline{\Gf})$, and $\FDTvar(\underline{\Pf_R}, \underline{\Gf})$ like so:
\begin{align*}
  \Act(\Val{lesion}, s, r) \coloneqq s \\
  \Act(\Val{nolesion}, s, r) \coloneqq r
\end{align*}
This is enough information to define $\Gf$ for the smoking lesion problem, which is given in \Fig{lesion.fdt}, and captures the fact that Fiona's action could depend on one of two different procedures, depending on whether or not she was born with the lesion.

We can define $\Pf_S$ on \Var{Luck}, \Var{Cancer}, \Var{Dead}, \Var{Outcome} and \VV analogously to Newcomb's problem. To finish defining $\Pf_S$, we also need to place probabilities on \Var{Lesion}, $\FDTvar(\underline{\Pf_R}, \underline{\Gf})$, and $\FDTvar(\underline{\Pf_S}, \underline{\Gf})$. For the first, let $\Pf_S(\Val{lesion})=p$ for some $0 < p < 1$ (for otherwise there's no uncertainty about the lesion). For the second, let $\Pf_S(\FDTvar(\underline{\Pf_R}, \underline{\Gf})=\Val{smoke})=q$ for some arbitrary $q$. (The correct $q$ is~0, as you may verify, but it won't matter if Fiona has uncertainty about how she'd act if her preferences were $\Util_R$ instead of $\Util_S$.) For the third, the prior probability is inconsequential, as we will soon see.

We are now ready to evaluate $\FDT(\Pf_S, \Gf)$ and thereby figure out what Fiona does in the case where her preferences are $\Util_S$. To calculate the expected utility of her options, Fiona constructs two hypotheticals. In the first, the variable $\FDTvar(\underline{\Pf_S}, \underline{\Gf})$ is set to \Val{smoke} by the $\Do$ operator; in the second, it is set to \Val{refrain}. In both cases, changing $\FDTvar(\underline{\Pf_S}, \underline{\Gf})$ does not affect the probability of \Val{lesion}:
\begin{align*}
  \Pf_S(\Val{lesion})\ &=\Pf_S(\Val{lesion}\mid\Do(\FDTvar(\underline{\Pf_S}, \underline{\Gf})=\Val{smoke}))\\
  & =\Pf_S(\Val{lesion}\mid\Do(\FDTvar(\underline{\Pf_S}, \underline{\Gf})=\Val{refrain})).
\end{align*}
This is just $p$. In English, this equation says that according to Fiona's hypotheticals, changing $\FDT(\Pf_S, \Gf)$ does not affect the probability that she has the lesion. She evaluates each hypothetical using $\Util_S$ (though she is not explicitly aware that she is using $\Util_S$ rather than $\Util_R$) and concludes that the \Val{smoke} hypothetical has expected utility
\[
  p(\text{\$10,000} + \text{\$1,000}) + (1-p)(\text{\$990,000} + q\text{\$1,000}),
\]
and the \Val{refrain} one has expected utility
\[
  p(\text{\$10,000}) + (1-p)(\text{\$990,000} + q\text{\$1,000}).
\]
She concludes that smoking is strictly better than refraining, by an amount equal to $p \cdot \text{\$1,000}$, corresponding to the probability that $\FDT(\Pf_S, \Gf)$ is actually the procedure controlling her action, times the value of smoking. Therefore, she smokes.

Thus we see that a single compact decision criterion, the functional decision theory given by \Eqn{fdt}, prescribes both one-boxing and smoking.

\citet{Eells:1984} has objected that when one fully formalizes this problem, as we have, it is revealed to be quite contrived---the agents are asked to make their choice under uncertainty about their own desires, which seems unrealistic. We agree, and note that the problem is more unfair to EDT than is generally supposed: Eve's distribution $\Pe$ lies to her! If Eve lives in a population of EDT agents, then \emph{none of them smoke}, and so there should be no correlation between \Act and \Var{Lesion}. To make $\Pe{}$ accurate, we would need Eve to live in a population of agents that use a different decision procedure than she does, while also demanding that Eve be ignorant of this fact (as well as her own desires).

The discussion above reveals that the smoking lesion problem is badly confused. We treat it in this \paper only because of its historical significance. Given that CDT and FDT handle it well despite its difficulties, and EDT can both handle it using a tickle defense and rightfully claim abuse, we recommend that philosophers abandon the dilemma. In \Sec{edt}, we will discuss alternative dilemmas that are less contrived, and which raise the same underlying issues for Eve in a manner that cannot be addressed by ratification or a tickle defense.

\subsection{Transparent Newcomb Problem}

\begin{figure}
    \centering
    \begin{subfigure}{\textwidth}
      \centering
      \subcaptionbox{Carl's graph $\Gc{}$ for the transparent Newcomb problem.\label{fig:tnewcomb.graph}}[.45\linewidth]{
        \begin{tikzpicture}[scale=0.8, every node/.style={transform shape}]
          \node[rounded rectangle,draw, minimum size=1cm] (nState)   at (3, 7.5) {\Var{Predisposition}};
          \node[rectangle,draw,minimum size=1cm] (nAct)   at (1, 3.5) {\Act};
          \node[rounded rectangle,draw,minimum size=1cm] (nPrediction) at (5, 5) {\Var{Prediction}};
          \node[rounded rectangle,draw,minimum size=1cm] (nNoise) at (6.25, 6.25) {\Var{Accurate}};
          \node[rounded rectangle,draw,minimum size=1cm] (nBoxB) at (5, 3.5) {\Var{Box B}};
          \node[rounded rectangle,double,draw,minimum size=1cm] (nObs) at (3, 3.5) {\Obs};
          \node[diamond,draw,minimum size=1cm] (nUtil)   at (3, 2) {\VV};
          \draw[->, >=latex] (nState) to[out=210,in=90] (nAct);
          \draw[->, >=latex] (nState) to[out=-30,in=90] (nPrediction);
          \draw[->, >=latex] (nNoise) -> (nPrediction);
          \draw[->, >=latex] (nPrediction) -> (nBoxB);
          \draw[->, >=latex] (nAct) -> (nUtil);
          \draw[->, >=latex] (nBoxB) -> (nUtil);
          \draw[->, >=latex] (nBoxB) -> (nObs);
          \draw[->, >=latex] (nObs) -> (nAct);
        \end{tikzpicture}}%
      \hfill
      \subcaptionbox{Fiona's graph $\Gf$ for the transparent Newcomb problem. If she sees the box full, she intervenes on \FDTpgv{full}, as illustrated. Otherwise, she intervenes on \FDTpgv{empty}.\label{fig:tnewcomb.fdt}}[.45\linewidth]{
        \begin{tikzpicture}[scale=0.8, every node/.style={transform shape}]
          \node[rectangle,draw, minimum size=1cm] (nState1)   at (5, 7.5)
          {\FDTpgv{full}};
          \node[rounded rectangle,draw, minimum size=1cm] (nState2)   at (2, 6.25) {\FDTpgv{empty}};
          \node[rounded rectangle,draw,minimum size=1cm] (nAct)   at (1.5, 3.5) {\Act};
          \node[rounded rectangle,draw,minimum size=1cm] (nNoise) at (6.25, 6.25) {\Var{Accurate}};
          \node[rounded rectangle,draw,minimum size=1cm] (nPrediction) at (5, 5) {\Var{Prediction}};
          \node[rounded rectangle,draw,minimum size=1cm] (nBoxB) at (5, 3.5) {\Var{Box B}};
          \node[diamond,draw,minimum size=1cm] (nUtil)   at (3, 2) {\VV};
          \draw[->, >=latex] (nState1) to[bend left=10] (nAct);
          \draw[->, >=latex] (nState1) -> (nPrediction);
          \draw[->, >=latex] (nState2) -> (nAct);
          \draw[->, >=latex] (nNoise) -> (nPrediction);
          \draw[->, >=latex] (nPrediction) -> (nBoxB);
          \draw[->, >=latex] (nPrediction) -> (nAct);
          \draw[->, >=latex] (nAct) -> (nUtil);
          \draw[->, >=latex] (nBoxB) -> (nUtil);
        \end{tikzpicture}}%
    \end{subfigure}
\end{figure}

To see how Eve and Carl fare in the transparent Newcomb problem, we need only make a small modification to their models for Newcomb's problem. In particular, we need to add a variable \Obs that reveals the contents of \Var{Box B}. Call the resulting probability distributions $\Pe$ and $\Pc$, and define $\Gc$ as in \Fig{tnewcomb.graph}. Now we simply calculate $\EDT(\Pe, \Val{full})$ and $\CDT(\Pc, \Gc, \Val{full})$ as per equations~\eq{edt} and~\eq{cdt}. The evaluation runs similarly to how it ran in Newcomb's problem above, except that the distributions are conditioned on $\Obs=\Val{full}$ before expected utility is calculated. Carl still two-boxes:
\begin{align*}
  & \EE(\VV \mid \Do(\Act=\Val{twobox}), \Obs=\Val{full}) = \text{\$1,001,000},\\
  & \EE(\VV \mid \Do(\Act=\Val{onebox}), \Obs=\Val{full}) = \text{\$1,000,000}.
\end{align*}
Eve also two-boxes in this case (assuming she does not assign prior probability 0 to two-boxing):
\begin{align*}
  & \EE(\VV \mid \Act=\Val{twobox}, \Obs=\Val{full}) = \text{\$1,001,000},\\
  & \EE(\VV \mid \Act=\Val{onebox}, \Obs=\Val{full}) = \text{\$1,000,000}.
\end{align*}
As such, it is easy to see that Carl and Eve would both two-box in the transparent Newcomb problem, so the predictor will not fill their boxes, and they will see empty boxes and walk away poor.

For Fiona, however, the case is quite different. Fiona does not react to observations by conditioning her distribution. Rather, she reacts to observations by switching which variable she intervenes on, where the different variables stand for the output of the FDT procedure when evaluated with different observations as input. This is captured by Fiona's graph $\Gf$ for this decision problem, which is given in \Fig{tnewcomb.fdt}. \Act depends on \FDTpgv{full}, \FDTpgv{empty}, and \Var{Prediction}, with \Var{Prediction} determining which of the two \FDTvar{} variables controls \Act. This represents the fact that the predictor's prediction determines which observation the agent receives.

Given $\Gf$ and the associated probability distribution $\Pf$, we can evaluate $\FDT(\Pf, \Gf, \Val{full})$ to determine how Fiona behaves, which involves finding the $\act$ that maximizes
\[
  \EE(\VV \mid \Do(\FDTpgv{full}=\act, \Obs=\Val{full}).
\]

In the case of $\act=\Val{onebox}$, the value of \VV is determined by the value of \Var{Accurate}. If $\Var{Accurate}=\Val{accurate}$, then \Var{Prediction} will be \Val{1}, \Var{Box B} will be \Val{full}, and \Act will be $a=\Val{onebox}$, so \VV will be \$1,000,000. If instead $\Var{Accurate}=\Val{inaccurate}$, then \Var{Prediction} will be \Val{2}, \Var{Box B} will be \Val{empty}, and \Act will be determined by $\FDTpgv{empty}$ instead. Write $q$ for Fiona's prior probability that she two-boxes upon seeing the box empty; in this case, $\VV$ will be $q\cdot \text{\$1,000}$ in expectation. (The correct value of $q$ is~1, as you may verify, but in this case it doesn't matter if Fiona has some uncertainty about how she would act in that case.) Total expected utility in the case of $\act=\Val{onebox}$ is therefore $\text{\$990,000} + 0.01q \cdot \text{\$1,000}$, because $\Var{Accurate}=\Val{accurate}$ with probability 99\%.

As you may verify by a similar calculation, in the case of $\act=\Val{twobox}$, total expected utility is $0.01 \cdot \text{\$1,001,000} + 0.99q \cdot \text{\$1,000} = \text{\$110,00} + q\cdot\text{\$990}$.

The first hypothetical has higher expected utility, so Fiona takes one box. What's remarkable about this line of reasoning is that even in the case where Fiona has observed that box B is full, when she \emph{envisions} two-boxing, she envisions a scenario where she instead (with high probability) sees that the box is empty. In words, she reasons: ``The thoughts I'm currently thinking are the decision procedure that I run upon seeing a full box. This procedure is being predicted by the predictor, and (maybe) implemented by my body. If it outputs \Val{onebox}, the box is likely full and my brain implements this procedure so I take one box. If instead it outputs \Val{twobox}, the box is likely empty and my brain does not implement this procedure (because I will be shown an empty box). Thus, if this procedure outputs \Val{onebox} then I'm likely to keep \$1,000,000; whereas if it outputs \Val{twobox} I'm likely to get only \$1,000. Outputting \Val{onebox} leads to better outcomes, so this decision procedure hereby outputs \Val{onebox}.''

The predictor, following the above chain of reasoning, knows that Fiona will one-box upon seeing the box full, and fills box B. Fiona sees two full boxes, takes only box B, and walks away rich.

\section{Diagnosing EDT: Conditionals as Counterfactuals} \label{sec:edt}

We are now in a better position to say why following CDT and EDT tends to result in lower-utility outcomes than following FDT: the hypotheticals that CDT and EDT rely on are malformed. In the hypotheticals that Carl constructs for Newcomb's problem, \Act is treated as if it is uncorrelated with the prediction even though the predictor is known to be highly reliable. In the hypotheticals that Eve constructs for the smoking lesion problem, \Var{Cancer} varies with \Act in spite of the fact that the correlation between them is merely statistical. EDT's hypotheticals respect too many correlations between variables; CDT's hypotheticals respect too few.

When an EDT agent imagines behaving a certain away, she imagines that all of the correlations between her action and the environment persist---even where there is no mechanism underlying the persistence. This is what leads EDT agents to irrationally ``manage the news,'' as \citet{Lewis:1981} puts it.

This phenomenon is easiest to see in cases where the EDT agent is certain that she will take a particular action \citep{Spohn:1977}. Consider, for example, a simple dilemma where an agent has to choose whether to take \$1 or \$100. If Eve is certain that she's going to take \$1, then she literally cannot imagine taking \$100---EDT agents condition, they don't counterfact. Thus, she takes \$1.

The standard defense of EDT here is that it's unrealistic to imagine Eve being completely certain about what action she's going to take. At the very least, Eve should allow that a wayward cosmic ray could strike her brain at a key moment and cause her to take the \$100. However, adding uncertainty does not fix the core problem. Consider:

\begin{dilemma}[The Cosmic Ray Problem] \label{dil:ray}
  An agent must choose whether to take \$1 or \$100. With vanishingly small probability, a cosmic ray will cause her to do the opposite of what she would have done otherwise. If she learns that she has been affected by a cosmic ray in this way, she will need to go to the hospital and pay \$1,000 for a check-up. Should she take the \$1, or the \$100?
\end{dilemma}
\noindent This hardly seems like it should be a difficult dilemma, but it proves quite troublesome for Eve. If, according to Eve's world-model, she almost always takes \$1, then it must be the case that whenever she takes \$100, it's because she's been hit by cosmic rays. Taking \$100 will then mean that she needs to go to the hospital, at a cost of \$1,000. Knowing this, Eve takes the \$1, for fear of cosmic rays. And her fears are (therefore) correct! She only \emph{does} take the \$100 when she's been hit by cosmic rays, and whenever that happens, she really does lose \$900 on net.

EDT runs into trouble because its hypotheticals do not allow Eve to consider breaking any correlations that hold between her action and the world. CDT and FDT agents don't fall into the same trap. If you tell Carl or Fiona that (by pure statistical happenstance) they only take \$100 when hit by cosmic rays, then they will dismiss your warning and take the \$100, and you will be revealed to be a liar. Carl and Fiona ignore (and therefore break) inconvenient statistical correlations between their actions and the environment. Eve lacks that capacity, to her detriment.

This failure makes EDT agents systematically exploitable. Consider the following dilemma, due to \citet{Soares:2015:toward}.

\begin{dilemma}[The XOR Blackmail] \label{dil:eblackmail}
  An agent has been alerted to a rumor that her house has a terrible termite infestation that would cost her \$1,000,000 in damages. She doesn't know whether this rumor is true. A greedy predictor with a strong reputation for honesty learns whether or not it's true, and drafts a letter:
\begin{quote}
I know whether or not you have termites, and I have sent you this letter iff exactly one of the following is true: (i) the rumor is false, and you are going to pay me \$1,000 upon receiving this letter; or (ii) the rumor is true, and you will \emph{not} pay me upon receiving this letter.
\end{quote}
  The predictor then predicts what the agent would do upon receiving the letter, and sends the agent the letter iff exactly one of (i) or (ii) is true.\footnote{To simplify exposition, we will assume here that the predictor is infallible. It's a trivial exercise to swap in a fallible predictor, similar to the predictors we've considered in previous dilemmas.} Thus, the claim made by the letter is true. Assume the agent receives the letter. Should she pay up?
\end{dilemma}

\noindent The rational move is to refuse to pay. If the agent is the type of agent who pays, then the letter will always be good news, since it will only come when her house is termite-free. If she's the type of agent who refuses, then the letter will always be bad news, since it will only come when she does have termites. But either way, the letter doesn't affect whether she has termites.

Eve, however, responds to the letter by paying up. To evaluate paying, she conditions on $\Act=\Val{pay}$, which (given that she has received the letter) sends her probability on \Val{termites} to 0. To imagine a world in which she does not pay, she conditions her world-model on $\Act=\Val{refuse}$, which (given that she has received the letter) sends the probability of \Val{termites} to 1. She prefers the former hypothetical, and so she pays. It follows that any sufficiently competent predictor who knows Eve has a reliable way to extract money from her, simply by presenting her with cleverly-crafted items of good news.

The XOR blackmail problem differs from the smoking lesion problem on a few counts. First and foremost, this is a problem where the tickle defense of \citet{Eells:1984} doesn't help Eve in the slightest---she pays up even if she has perfect self-knowledge. She also can't be saved by ratification---she pays up even if she knows she is going to pay up. The only way to keep Eve from paying blackmailers in this scenario is to push her probability $\Pe{}(\Act=\Val{pay})$ to zero, so that she literally cannot imagine paying. If EDT agents can only reliably achieve good outcomes when we're kind enough to identify the best action ourselves and indoctrinate the agent into believing she'll choose that, then as \citet{Joyce:2007} points out, EDT cannot reasonably claim to be the correct theory of rational choice.

Carl and Fiona, unlike Eve, refuse to pay in \Dil{eblackmail}. It is worth noting, however, that they refuse for very different reasons.

Carl refuses because the variable \Var{Infestation} representing the infestation is causally upstream from \Act in his world-model (for which the graph is given in \Fig{eblackmail.cdt}). Write ${p=\Pc(\Var{Infestation}=\Val{termites})}$ for his prior probability that his house is infested. According to Carl's hypotheticals, $p$ does not change with \Act, so there isn't any point in paying.

\begin{figure}
    \centering
    \begin{subfigure}{\textwidth}
    \centering
      \subcaptionbox{A causal graph $\Gc{}$ for Carl in the XOR blackmail problem.\label{fig:eblackmail.cdt}}[.46\linewidth]{
      \begin{tikzpicture}[scale=0.8, every node/.style={transform shape}]
        \node[rounded rectangle,draw, minimum size=1cm] (nAsteroid) at (0.5, 6.5) {\Var{Infestation}};
        \node[rounded rectangle,draw,minimum size=1cm] (nPredictor) at (3, 5) {\Var{Predictor}};
        \node[rectangle,draw,minimum size=1cm] (nAct) at (3, 3.5) {\Act};
        \node[rounded rectangle,double,draw,minimum size=1cm] (nObs) at (4.5, 3.5) {\Obs};
        \node[diamond,draw,minimum size=1cm] (nUtil)   at (3, 2) {\VV};
        \draw[->, >=latex] (nAsteroid) to[out=-90,in=180] (nUtil);
        \draw[->, >=latex] (nAsteroid) -> (nPredictor);
        \draw[->, >=latex] (nPredictor) -> (nAct);
        \draw[->, >=latex] (nPredictor) -> (nObs);
        \draw[->, >=latex] (nAct) -> (nUtil);
      \end{tikzpicture}}    \hfill
      \subcaptionbox{A causal graph $\Gf$ for Fiona in the same problem, with the point of intervention illustrated as in the case where Fiona sees the letter.\label{fig:eblackmail.fdt}}[.46\linewidth]{
      \begin{tikzpicture}[scale=0.8, every node/.style={transform shape}]
        \node[rounded rectangle,draw, minimum size=1cm] (nAsteroid) at (0.5, 6.5) {\Var{Infestation}};
        \node[rectangle,draw,minimum size=1cm] (nAlgo) at (4.5, 6.5) {\FDTpgv{letter}};
        \node[rounded rectangle,draw,minimum size=1cm] (nAlgo2) at (5, 3.5) {\FDTpgv{$\emptyset$}};
        \node[rounded rectangle,draw,minimum size=1cm] (nPredictor) at (2.5, 5) {\Var{Predictor}};
        \node[rounded rectangle,draw,minimum size=1cm] (nAct) at (2.5, 3.5) {\Act};
        \node[diamond,draw,minimum size=1cm] (nUtil)   at (3, 2) {\VV};
        \draw[->, >=latex] (nAsteroid) to[out=-90,in=180] (nUtil);
        \draw[->, >=latex] (nAsteroid) -> (nPredictor);
        \draw[->, >=latex] (nAlgo) -> (nPredictor);
        \draw[->, >=latex] (nAlgo) to[bend left=25] (nAct);
        \draw[->, >=latex] (nAlgo2) -> (nAct);
        \draw[->, >=latex] (nPredictor) -> (nAct);
        \draw[->, >=latex] (nAct) -> (nUtil);
      \end{tikzpicture}}
    \end{subfigure}
    \label{fig:eblackmail}
\end{figure}

What about Fiona? Write $(\Pf, \Gf)$ for her world-model, with $\Gf$ given by \Fig{eblackmail.fdt} and $\Pf$ defined in the obvious way. From $\Gf$, we can see that according to Fiona's hypotheticals, changing $\FDT(\Pf, \Gf, \Val{letter})$ changes what Fiona would do if she received the letter, \emph{and} changes the probability that she sees the letter in the first place. It does not, however, change her probability that her house is infested; so she refuses to pay (as you may verify).

The actions of Carl and Fiona are the same in this dilemma, and their beliefs about the actual world are similar, but their \emph{hypotheticals} are very different. If Fiona has termites and sees the letter, and you ask her what would have happened if she had paid, she will consult her $\Do(\FDTpgv{letter}=\Val{pay})$ hypothetical, and tell you that in that case the predictor would have predicted differently and would not have sent the letter, so she would have been running $\FDT(\Pf, \Gf, \emptyset)$ instead. By contrast, if you ask Carl what would have happened if he had paid up, he will consult his $\Do(\Act=\Val{pay})$ hypothetical, and report back that he would still have gotten the letter, the house would still have been infested, and he would have lost money. In other words, Carl's hypothetical would say that the predictor erred---despite the fact that the predictor is inerrant, \emph{ex hypothesi}. This hints at the reason why CDT fails in other dilemmas.

\section{Diagnosing CDT: Impossible Interventions} \label{sec:cdt}

EDT agents fail because they cannot imagine breaking correlations between their action and their environment. Meanwhile, CDT agents fail because they imagine breaking \emph{too many} correlations between action and environment.

Imagine Carl facing Newcomb's problem against an absolutely perfect predictor. We confront Carl as he walks away, a paltry \$1,000 in hand, and ask him what would have happened if he had taken one box. To answer, he will use CDT hypotheticals. He will envision his action changing, and he will envision everything that depends causally on his action changing with it, while all else is held constant. As such, he will envision a scenario in which box B stays empty, but box A is left behind. He will say: ``Then I would have left empty-handed; I sure am glad I two-boxed.''

Contrast this with what \emph{we} know would have happened if Carl had one-boxed (if CDT were the predictably-onebox sort of decision theory). Box B would have been full, and Carl would have been rich.

Carl answers the ``what if?'' question with a description of a hypothetical scenario in which a perfect predictor made an incorrect prediction. Carl's response to the XOR blackmail problem was similarly odd. This is the flaw in CDT hypotheticals: CDT agents imagine that their actions are uncorrelated with the behavior of environmental processes that implement the same decision procedure. According to Carl's hypotheticals, known-to-be-perfect predictors inexplicably falter exactly when Carl makes his current decision. It is no wonder, then, that Carl runs into trouble in settings where the environment contains predictors or twins.

This flaw is more serious than is generally recognized. To see this, we will consider a rather cruel dilemma in which the predictor punishes all agents no matter what they choose, inspired by the ``Death in Damascus'' dilemma popularized by \citet{Gibbard:1978}:

\begin{quote}
Consider the story of the man who met Death in Damascus. Death looked surprised, but then recovered his ghastly composure and said, `\textsc{i am coming for you tomorrow.}' The terrified man that night bought a camel and rode to Aleppo. The next day, Death knocked on the door of the room where he was hiding, and said `\textsc{i have come for you.}'

`But I thought you would be looking for me in Damascus,' said the man.
`\textsc{not at all},' said Death `\textsc{that is why i was surprised to see you yesterday. i knew that today i was to find you in aleppo.}'
\end{quote}
\noindent There are many different ways to turn this story into a decision problem. We choose the simplest version, copying Gibbard and Harper's structure. We add a cost of \$1,000 if the agent decides to flee Damascus, reasoning that most people prefer to spend their final night with their loved ones rather than on camelback.
\begin{dilemma}[Death in Damascus] \label{dil:death}
Imagine a deterministic world where Death is known to be able to perfectly predict human behavior based on a detailed past observation of the world's state. Death works from an appointment book, which lists combinations of people, days, and places. Each day, Death goes to collect the listed people at the listed places. If the listed person is at the corresponding place on that day, they die; otherwise, they survive (which they value at \$1,000,000).

An agent encounters Death in Damascus and is told that Death is coming for her tomorrow. This agent knows that deciding to flee to Aleppo (at a cost of \$1,000) means that Death will be in Aleppo tomorrow, whereas staying in Damascus means that Death will be in Damascus tomorrow. Should she stay, or flee?
\end{dilemma}
\noindent The correct decision here is rather obvious. The \$1,000,000 is a lost cause, but the agent can save \$1,000 by staying in Damascus. Fiona recognizes this, and concludes that wherever she goes, Death will be. She doesn't waste her final night fleeing.

CDT agents, in contrast, are sent into \emph{conniptions} by this dilemma. As Gibbard and Harper put it, CDT in this situation is ``unstable.'' Carl bases his decisions on hypotheticals in which Death's action is independent of his own action. This means that if he initially expects himself to stay in Damascus, then he will want to go to Aleppo---as though he expects Death to perfectly predict all of his behavior \emph{except} for this final decision. By the same token, if he initially expects himself to flee to Aleppo, then he will want to stay in Damascus after all.

To formalize this line of thinking, we'll let $\Pc_1$ be the initial belief state of Carl when he faces the Death in Damascus dilemma. (We leave it to the reader to draw Carl's graph $\Gc$ for this problem.) As noted by Gibbard and Harper, whatever epistemic state Carl occupies will be reflectively inconsistent. Assume that Carl thinks he is going to stay in Damascus, i.e., that $\Pc_1(\Val{damascus}) > \Pc_1(\Val{aleppo})$. Then he will go to Aleppo, i.e., $\CDT(\Pc_1, \Gc)=\Val{aleppo}$. Assuming Carl is smart enough to recognize his new travel plans, he now needs to update his beliefs to $\Pc_2$ where $\Pc_2(\Val{aleppo}) \approx 1$. But then $\CDT(\Pc_2, \Gc)=\Val{damascus}$, so now Carl needs to update his beliefs again!

There are four ways for Carl to break out of this loop. First, he could hold to the delusion that he has chosen Damascus even as he rides to Aleppo (or vice versa). Second, he could sit there updating his beliefs all day long, constructing $\Pc_3$ and $\Pc_4$ and $\Pc_5$ and never solving \Eqn{cdt} for any of them, until Death comes by to pick him up in the morning. Third, he could give up after $n$ iterations and follow $\CDT(\Pc_n, \Gc)$ to either Damascus or Aleppo, resigned to failure. Or fourth, he could use a source of pseudorandomness that he himself cannot predict (but which Death can predict, \emph{ex hypothesi}), and get himself into a ``ratified'' belief state $\Pc_R$ where he is uncertain about what he's going to do, but which is consistent under reflection, as suggested by \citet{Arntzenius:2008}.

\citet{Joyce:2012} describes and endorses a ratification procedure in keeping with the fourth option. Briefly, this procedure leaves Carl in a state of knowledge where he is indifferent between all actions that have positive subjective probability, at those probabilities. In this case, he ends up assigning 50.05\% probability to \Val{damascus} and 49.95\% probability to \Val{aleppo}, because that is the point where the extra utility he expects to receive from staying in Damascus exactly balances out his extra subjective probability that Death will be there.

Regardless, if and when Carl finally does pick a belief state $\Pc_\ast$ and run $\CDT(\Pc_\ast, \Gc)$, it will have him weigh his options by imagining hypothetical scenarios in which Death's location is uncorrelated with his action.\footnote{If Carl uses the ratification procedure of \citet{Joyce:2012}, then he actually finds the \emph{set} of actions $a$ which maximize $\EE(\VV \mid \Do(\act))$, and uses his pseudorandomness source to sample from among those, with probabilities proportional to $\Pc_\star(\act)$---i.e., he will use a source of pseudorandomness to go to Aleppo 49.95\% of the time and Damascus 50.05\% of the time, at which point he will die with certainty.}

There are two serious problems with Carl's reasoning here, if the aim is to end up with as much utility as possible. First, with $\approx$50\% probability, Carl will flee to Aleppo and lose the equivalent of \$1,000 in easy money. Second, Carl will \emph{behave as if} he has at least a 50\% subjective probability of survival.

To see why the latter is a serious problem, let us suppose that we wait until Carl finishes settling on some belief, and we then offer him (for the price of \$1) a coin that is \emph{truly} random---in the sense that if he bases his decision on a toss of this coin, he has a 50\% chance of thwarting Death.

Clearly, if Carl never breaks his infinite loop, he won't be able to take advantage of the coin. If he instead gives in to delusion, or if he gives up after $n$ iterations, then he now has some subjective probability $p$ (the exact value won't matter) that he is going to stay in Damascus. Because he knows that Death will be where he is, he also assigns $p$ probability to Death being in Damascus. But he \emph{acts as if} these variables are independent, which means that when he imagines staying in Damascus, he calculates expected utility \emph{as if} there were an \emph{independent} chance $q$ of Death being in Damscus, with $q$ ``coincidentally'' equal to $p$. Similarly, when he imagines fleeing to Aleppo, he calculates expected utility as if there is an independent $(1-q)$ chance of Death being there there. Carl therefore acts as if his chances of survival are $\max(q, 1-q) \ge 0.5$. By contrast, when he imagines buying your truly random coin, he calculates that it would give him a 50\% chance of survival---which, according to CDT's hypotheticals, isn't worth the dollar. ``No, thanks,'' Carl says. ``After thinking about my options, I've decided that I'd rather take my chances with certain death.''

CDT with pseudorandomization \`{a} la \citet{Arntzenius:2008} or \citet{Joyce:2012} performs no better, although we omit the technical details here. \mkbibparens{The matter is discussed in more depth by \citet{Ahmed:2014b}.} In Joyce's model, Carl chooses as if his pseudorandomness source were uncorrelated with Death's location, and thus he calculates utility as if tossing his pseudorandom coin gets him \$500,500 in expectation. This is exactly the utility he would expect to get by tossing the truly random coin, so he still concludes that the coin isn't worth a dollar.

This behavior is \emph{wildly} irrational. It's one thing to fail to recognize when resistance is futile; it's quite another to pass up a 50\% chance of an outcome worth \$1,000,000 at a cost of \$1. \citet{Ahmed:2014} has raised similar concerns. In fact, it is possible to leverage these irrationalities to turn CDT into a money-pump. Developing that machinery is outside the scope of this \paper; our focus here is on \emph{why} CDT goes wrong, and how FDT avoids this species of error.

Carl never \emph{believes} that he can escape Death. He knows that Death is a perfect predictor, and that he is doomed no matter what he does. And Carl is no happier to waste his final hours fleeing to Aleppo than Fiona would be. The reason why Carl fails where Fiona succeeds is in the hypotheticals that he consults when he weighs his options.

When it comes to beliefs of fact, Carl recognizes that Death will meet him wherever he goes. He can easily predict that, conditional on fleeing to Aleppo, he will die in Aleppo, and vice versa. He knows that Death can predict him even if he uses a source of pseudorandomness, and he can predict that, conditional upon using the truly random coin, he has a 50\% chance of survival. But he does not base his decisions on conditionals. He always and only chooses the action that corresponds to the highest expected utility \emph{in a CDT hypothetical}---and according to CDT, he should rely on hypotheticals in which Death's location isn't correlated with his action. According to \emph{those} hypotheticals, the coin is worthless. This mismatch between Carl's beliefs and his decision-theoretic hypotheticals provides further reason to suspect that CDT is not the correct theory of rational choice.

Causal decision theorists have insisted that the Death in Damascus problem is one where no matter what action the agent takes, it will turn out to have been the wrong one. If they stay in Damascus, they should have fled to Aleppo; if they flee, they should have stayed \citep{Gibbard:1978}. In their hypotheticals, they imagine their location changing while Death's location remains the same. Upon finding Death in Aleppo, they therefore reason that they ``could have'' survived had they but stayed in Damascus; or if they encounter Death in Damascus, they feel they ``could have'' survived by fleeing to Aleppo.

Functional decision theorists imagine the situation differently. FDT agents observe that Death's location causally depends on an accurate prediction in Death's book, which was made via an accurate prediction of the agent's decision procedure. When they imagine having made a different choice, they imagine a world in which Death's book has different writing in it. In this way their imagined hypotheticals avoid tossing out decision-relevant information about the world.

For that reason, we disagree with the analysis of \citeauthor{Gibbard:1978} when they say that ``any reason the doomed man has for thinking he will go to Aleppo is a reason for thinking he would live longer if he stayed in Damascus, and any reason he has for thinking he will stay in Damascus is reason for thinking he would live longer if he went to Aleppo. Thinking he will do one is reason for doing the other.'' This instability only arises if we imagine changing our decision without imagining Death's prediction changing to match---a strange thing to imagine, if the task at hand is to rationally respond to an environment containing accurate predictors.

When Fiona faces this dilemma, she quickly accepts that if she flees, Death will meet her in Aleppo. The reason she ends up with greater utility than Carl is that the mental operations she performs to construct her hypotheticals track the real-world dependence relations that she believes exist---the correspondence between predictors and the things they predict, for example. CDT hypotheticals neglect some of these dependencies, and Carl pays the price.

\section{The Global Perspective} \label{sec:perspective}

\citet{Gibbard:1978} and \citet{Lewis:1981} have argued that Newcomblike problems unfairly punish rational behavior. CDT agents cannot help being CDT agents; and just as we can construct dilemmas that doom CDT or EDT agents, so we can construct dilemmas that doom FDT agents. Consider dilemmas in which a mind reader gives \$1,000,000 to an agent iff she predicts they would two-box in Newcomb's problem, or iff she predicts they follow EDT.

We grant that it is possible to punish agents for using a specific decision procedure, or to design one decision problem that punishes an agent for rational behavior in a different decision problem. In those cases, no decision theory is safe. CDT performs worse that FDT in the decision problem where agents are punished for using CDT, but that hardly tells us which theory is better for \emph{making decisions}. Similarly, CDT performs poorly in the game ``punish everyone who takes \$100 in the cosmic ray problem,'' but this doesn't constitute evidence against CDT. No one decision theory outperforms all others in all settings---dominance of that form is impossible.

Yet FDT does appear to be superior to CDT and EDT in all dilemmas where the agent's beliefs are accurate and the outcome depends only on the agent's behavior in the dilemma at hand. Informally, we call these sorts of problems ``fair problems.'' By this standard, Newcomb's problem is fair; Newcomb's predictor punishes and rewards agents only based on their actions. If the predictor scanned Carl's brain, deduced that he followed causal decision theory, and punished him \emph{on those grounds}, then functional decision theorists would agree that Carl was being unfairly treated. But the predictor does no such thing; she merely predicts whether or not Carl will take one box, and responds accordingly. Or, in sound-bite terms: Newcomb's predictor doesn't punish rational agents; she punishes \emph{two-boxers}. She doesn't care how or why you one-box, so long as you one-box predictably.

There is no perfect decision theory for all possible scenarios, but there may be a general-purpose decision theory that matches or outperforms all rivals in fair dilemmas, if a satisfactory notion of ``fairness'' can be formalized.\footnote{There are some immediate technical obstacles to precisely articulating this notion of fairness. Imagine I have a copy of Fiona, and I punish anyone who takes the same action as the copy. Fiona will always lose at this game, whereas Carl and Eve might win. Intuitively, this problem is unfair to Fiona, and we should compare her performance to Carl's not on the ``act differently from Fiona'' game, but on the analogous ``act differently from Carl'' game. It remains unclear how to transform a problem that's unfair to one decision theory into an analogous one that is unfair to a different one (if an analog exists) in a reasonably principled and general way.} FDT's initial success on this front seems promising, and the authors have not been able to construct an intuitively fair dilemma where FDT loses to a rival decision theory.

As a final illustration of FDT's ability to outperform CDT and EDT simultaneously in surprising ways, consider a more general blackmail problem described by \citet{Soares:2015:toward}:

\begin{dilemma}[The Mechanical Blackmail] \label{dil:cblackmail}
  A blackmailer has a nasty piece of information which incriminates both the blackmailer and the agent. She has written a computer program which, if run, will publish it on the internet, costing \$1,000,000 in damages to both of them. If the program is run, the only way it can be stopped is for the agent to wire the blackmailer \$1,000 within 24 hours---the blackmailer will not be able to stop the program once it is running. The blackmailer would like the \$1,000, but doesn't want to risk incriminating herself, so she only runs the program if she is quite sure that the agent will pay up. She is also a perfect predictor of the agent, and she runs the program (which, when run, automatically notifies her via a blackmail letter) iff she predicts that she would pay upon receiving the blackmail. Imagine that the agent receives the blackmail letter. Should she wire \$1,000 to the blackmailer?
\end{dilemma}

\noindent Eve and Carl would pay up, of course. Given that the program is already running, they reason that they must choose between losing \$1,000 to the blackmailer or \$1,000,000 when the news gets out. In that light, paying up is clearly the lesser of the two evils. The blackmailer, knowing that Carl and Eve reason in this fashion, knows that it is safe to blackmail them. In this way, CDT and EDT give blackmailers incentives to exploit agents that follow the prescriptions of those theories.

If Eve or Carl had been the sort of agents who refuse to pay in this setting, then they would have been better off---and indeed, Carl and Eve would both pay for an opportunity to make a binding precommitment to refuse payment. What about Fiona?

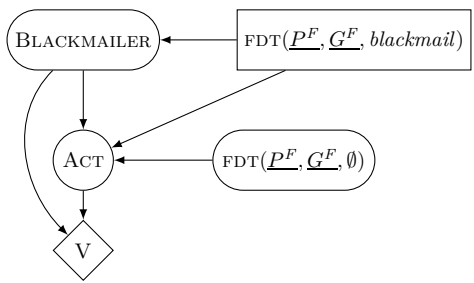
\begin{figure}
    \centering
    \begin{tikzpicture}[scale=0.8, every node/.style={transform shape}]
      \node[rectangle,draw,minimum size=1cm] (nAlgo) at (7.5, 5.5) {\FDTpgv{blackmail}};
      \node[rounded rectangle,draw,minimum size=1cm] (nAlgo2) at (6.5, 3.5) {\FDTpgv{$\emptyset$}};
      \node[rounded rectangle,draw,minimum size=1cm] (nPredictor) at (3, 5.5) {\Var{Blackmailer}};
      \node[rounded rectangle,draw,minimum size=1cm] (nAct) at (3, 3.5) {\Act};
      \node[diamond,draw,minimum size=1cm] (nUtil)   at (3, 2) {\VV};
      \draw[->, >=latex] (nPredictor) to[out=-135,in=135] (nUtil);
      \draw[->, >=latex] (nAlgo) -> (nPredictor);
      \draw[->, >=latex] (nAlgo) -> (nAct);
      \draw[->, >=latex] (nAlgo2) -> (nAct);
      \draw[->, >=latex] (nPredictor) -> (nAct);
      \draw[->, >=latex] (nAct) -> (nUtil);
    \end{tikzpicture}
    \caption{A graph $\Gf$ for Fiona in the mechanical blackmail problem. The point of intervention is drawn for the case where she observes the blackmail. \Var{Blackmailer} determines whether \FDTpgv{blackmail} or \FDTpgv{$\emptyset$} gets to set \Act.}
    \label{fig:cblackmail}
\end{figure}

In the case where Fiona \emph{has} received the message, Fiona refuses to pay up, for the same reason that she takes one box in the transparent Newcomb problem. Write $(\Pf, \Gf)$ for Fiona's world-model when facing a mechanical blackmail. $\Gf$ is given in \Fig{cblackmail}; $\Pf$ is defined in the obvious way. Upon receiving the blackmail, Fiona solves the equation $\FDT(\Pf, \Gf, \Val{blackmail})$, which requires consulting two hypotheticals, one in which \FDTpgv{blackmail} is set (by intervention) to \Val{pay}, and one where it is set to \Val{refuse}. In the first one, her probability of losing \$1,000,000 if blackmailed goes to $\approx$0 and her probability of being blackmailed goes to $\approx$1. In the second one, her probability of losing \$1,000,000 if blackmailed goes to $\approx$1, but her probability of being blackmailed goes to~$\approx$0!

According to FDT, when Fiona sees the blackmail and imagines refusing to pay, she should imagine a hypothetical world in which FDT is the kind of decision theory that refuses---which means that she should imagine a world in which she was never blackmailed in the first place. Because she chooses her actions \emph{entirely} by comparing the hypotheticals that FDT says to consider, Fiona concludes that refusing to pay is better than paying. The blackmailer, recognizing this, does not attempt to blackmail Fiona, and Fiona walks away unscathed.

At first glance, it may seem like Fiona is ignoring information---shouldn't she update on the fact that she has seen the blackmail?
\todo[inline,color=yellow]{Footnote: This apparent ``lack of updating'' is the source of the name of Wei Dai's ``Updateless Decision Theory'' \citep{Dai:2009}, an early variant of functional decision theory. Another early variant, the ``Timeless Decision Theory'' of \citet{Yudkowsky:2010:TDT}, prescribes both conditioning on evidence \emph{and} changing the place where one intervenes; this creates a kind of ``double update'' which proves harmful in practice.}{}
Fiona does change her behavior in response to evidence, though---by changing \emph{the place in her graph where she intervenes}. If she had not been blackmailed, she would have intervened on a different node \FDTpgv{$\emptyset$} representing her belief state given that she was not blackmailed. FDT says that agents should entertain the hypothesis that, if their action were different, they might have \emph{made different observations}.\footnote{As mentioned earlier, the author's preferred formulation of FDT actually intervenes on the node $\FDT(-)$ to choose not an action but a \emph{policy} which maps inputs to actions, to which the agent then applies her inputs in order to select an action. The difference only matters in multi-agent dilemmas so far as we can tell, so we have set that distinction aside in this \paper for ease of exposition.}

If Fiona \emph{did} receive the blackmail, she \emph{would} reason, ``Paying corresponds to a world where I lose \$1,000; refusing corresponds to a world where I never get blackmailed. The latter looks better, so I refuse.'' As such, she never gets blackmailed---her counterfactual reasoning is proven correct.

Fiona plays the same strategy even if the blackmailer is an imperfect predictor. Assume that with probability 0.0009, the blackmailer wrongly predicts that an agent will pay. Fiona therefore imagines that if she pays upon receiving blackmail then she always gets blackmailed and always loses \$1,000, whereas if she refuses to pay then she is blackmailed 0.09\% of the time and loses \$900 in expectation. As such, she refuses to pay, even when holding the blackmail letter, even though she \emph{knows} that the blackmailer predicted wrongly---because it's advantageous to be \emph{the kind of agent} that makes such decisions.

If instead the chance of error is 0.011\%, then she pays, as this is again the action that tends to make an agent rich. When weighing actions, Fiona simply imagines hypotheticals corresponding to those actions, and takes the action that corresponds to the hypothetical with higher expected utility---even if that means imagining worlds in which her observations were different, and even if that means achieving low utility in the world corresponding to her actual observations.

Upon holding the blackmail letter, would Fiona have a desire to self-modify and temporarily follow CDT? No! For according to her hypotheticals, if FDT was the sort of algorithm that connived to pay up upon receiving blackmail, it would be a tempting blackmail target, and she would be blackmailed more often, and tend to lose utility on net. Functional decision theory is, so far as we know, reflectively stable in fair problems.

CDT agents imagine hypothetical worlds in which, if only they had acted differently, they would have thwarted all predictions. FDT agents imagine hypothetical words in which, if their action were different, they would have seen different things. Both hypotheticals contain impossibilities---CDT agents always \emph{actually} end up in worlds where the predictions were accurate, and FDT agents always \emph{actually} end up in worlds where their observations are consistent with their actions. The only difference is a practical one: while Carl is stuck complaining that predictors ``punish rationality,'' Fiona is thwarting blackmailers and getting rich.

Still, we expect some lingering resistance to the notion that one should (in real life!) refuse to pay the blackmailer upon receiving a mechanical blackmail letter, or refrain from two-boxing upon observing two full boxes in the transparent Newcomb problem. These are odd conclusions. It might even be argued that sufficiently odd behavior provides evidence that what FDT agents see as ``rational'' diverges from what humans see as ``rational.'' And given enough divergence of that sort, we might be justified in predicting that FDT will systematically fail to get the most utility in some as-yet-unknown fair test.

One way of giving teeth to the notion that FDT is ``odd'' might be to argue that FDT agents have mistaken views of causality. Fiona acts as though she can control abstract mathematical functions and events that happened in her past. On this view, the dilemmas we have discussed in this paper reward FDT's particular delusions, but they are delusions. We should therefore be skeptical that FDT's utility-reaping behavior in the hand-picked dilemmas in this paper are reflective of utility-reaping behavior across the board.

We respond that FDT agents suffer from no delusion; or, at least, they are no more deluded than their counterpart CDT agents. Consider CDT's prescriptions in deterministic dilemmas. The causal decision theorist argues that although there is only one action a deterministic physical agent can truly take, the sorts of agents that do well are the ones that \emph{imagine} hypothetical worlds across which their action differs, and take the action corresponding to the hypothetical with highest expected utility. The functional decision theorist is in a similar situation, arguing that although there is only one output a given function can have on a given input, an agent will do better if she \emph{imagines} hypothetical worlds across which the output of her decision function varies. Any principled objection to the idea that FDT agents can ``control'' their decision type will also imply that (deterministic) CDT agents cannot ``control'' their decision token. Since CDT and FDT face analogous difficulties in making sense of the idea of ``control,'' this factor cannot help us decide between the theories.

More generally, we have found that FDT's prescriptions look less odd when we think of decision-making as a natural phenomenon like any other, and require that good decisions take into account agents' beliefs and preferences about the world as a whole (subjunctive dependencies included), not just their beliefs and preferences about events causally downstream from the decision. In a sense, FDT's prescription is that an agent should choose her action from a global perspective---asking herself ``What would the entire universe look like if the FDT algorithm behaved differently?'' and choosing the action that corresponds to the best imagined universe. If that means imagining a universe in which box B is empty (even though she can clearly see that it is full), or imagining a universe in which she was probably never blackmailed in the first place (even as she's holding the blackmail letter in her hands), then so be it.

In \Sec{edt}, we saw that EDT agents fail to imagine that they can break merely statistical correlations between their actions and their environment. This puts Eve at the mercy of anyone who can deliberately produce correlations between Eve's actions and items of bad news. In \Sec{cdt}, we saw that CDT agents fail to imagine that they can control logically necessary correlations between their actions and their environment---causing Carl to make erratic and self-defeating decisions in the presence of predictors. EDT respects too many action-environment correlations, while CDT respects too few.

FDT, we claim, gets the balance right. An agent who weighs her options by imagining worlds where her decision function has a different output, but where logical, mathematical, nomic, causal, etc. constraints are otherwise respected, is an agent with the optimal predisposition for whatever fair dilemma she encounters.

\section{Conclusion} \label{sec:conclusions}

Functional decision theory demonstrates that, contra \citet{Gibbard:1978}, a single general-purpose normative rule can prescribe one-boxing in Newcomb's problem and smoking in the smoking lesion problem. Unlike EDT agents, FDT agents don't manage the news or pay to avoid information. Unlike CDT agents, FDT agents can account for predictors and twins in stable, consistent, and utility-maximizing ways. And unlike both CDT agents and EDT agents, FDT agents will hitch a ride out of Parfit's desert. FDT agents avoid making decisions they will immediately regret, and avoid giving other agents an incentive to extort them.

FDT achieves all of this with one elegant criterion, similar in form to CDT. We know of no fair dilemmas in which FDT systematically achieves lower utility than a rival theory; and it succeeds without our needing to make any ad-hoc emendations to the theory to accommodate certain classes of dilemmas, and without any need for complicated ratification procedures like those of \citet{Jeffrey:1983}, \citet{Eells:1984}, or \citet{Joyce:2012}. Nor is there any need for FDT agents to adopt costly precommitment mechanisms: FDT agents always act as they would have precommitted to act.

Given a probabilistic model of the world and a theory of subjunctive dependencies saying how the universe would look different if (counterpossibly) the FDT function had different outputs, FDT yields a concrete step-by-step procedure that one can follow to reap these benefits, without committing oneself to any particular doctrine of free will, or the metaphysics of possible worlds, or what an agent ``really'' can and cannot do. We claim only that if an agent wants to get as much utility as she can, she should weigh different imaginary worlds in which her decision procedure has different logical outputs, and execute the action corresponding to the best such world.

Though we briefly outlined our interpretation of FDT as a naturalistic and ``global'' approach to decision-making, our primary case for FDT has simply been that it performs much better in various dilemmas. FDT works where CDT and EDT fail.

The fact that FDT works matters for real-world decision-making. Newcomblike problems are common in practice \citep{Lewis:1979}, and the issues underlying Newcomblike problems lie at the very heart of decision theory. This becomes more obvious when we consider decision-theoretic and game-theoretic dilemmas together: the basic problem of achieving mutual cooperation and group coordination, of making credible contracts and alliances while resisting threats and extortion, is a problem of reasoning in the face of (imperfect) predictors and other replicators of one's behavior.

Consider a dilemma in which an agent needs to decide whether to spend \$1,000 to access a voting booth, and will receive \$1,000,000 iff she and nine other agents all independently choose to vote. If these ten agents have common knowledge that everyone in the group follows CDT, then none of them will bother to vote (unless, for some reason, they are already convinced at the outset that the other nine will vote). A member of a group of FDT agents, in contrast, will recognize that the group members' decisions are not subjunctively independent. She will therefore vote---not out of altruism, but out of an awareness that the other members of the group will reason symmetrically and also vote.

Or consider dilemmas in which an agent needs to coordinate with herself over time. Perhaps the agent assigns \$1,000,000 in utility to getting fit (which requires going to the gym a few dozen times), but she hates working out and assign \$1,000 in utility to skipping her scheduled gym day. Carl the CDT agent has no way to force himself to stick to his guns, and will always avoid the gym, because he doesn't think his action today is necessary or sufficient for getting his future selves to work out. Fiona the FDT agent, in contrast, can see that the decision she faces this week is similar to the decision she will face next week. If there are no relevant differences between her decision today versus tomorrow, then she should assume that she will continue going to the gym in the future if she goes today; and she should assume that she will continue skipping the gym if she skips today. Since the former is preferable, she goes to the gym.

The distinction between FDT, CDT, and EDT is particularly essential in the domain of computer science. Computer programs are easier to copy than human brains, and their behavior is often easier to predict. As artificial intelligence systems become more common, we should expect them to frequently enter Newcomblike dilemmas with each other. If an AI system's programmers frequently base their behavior on predictions about how the system is going to behave, then they are likely to put the agent into Newcomblike dilemmas. If we want to avoid the kind of bizarre behavior EDT exhibits in the XOR blackmail problem or CDT exhibits in the twin prisoner's dilemma, we will need to formalize practical alternatives.

The consistent ability of FDT agents to achieve the best outcomes in fair dilemmas suggests that FDT may be the correct normative theory of rational choice. Neither CDT nor EDT wins unambiguously as measured by utility achieved in Newcomblike problems, so it was natural to fall back to the notion of dominance in attempt to distinguish between them. However, with FDT in hand, the principle of utility maximization is revived---which is good news, given the weakness of the principle of dominance that we discussed in \Sec{revisit}.

The adjustments required to get from CDT to FDT, however, are not small. FDT is a genuinely novel theory, and raises a number of new practical and philosophical questions. Chief among these is whether we can find some method for deducing mathematical, logical, etc. subjunctive dependencies from observation and experiment, analogous to Pearl's \citeyearpar{Pearl:2000} method for deducing causal dependencies. And while we have made an informal case for FDT's superiority by looking at particular well-known decision problems, we would prefer to have a general-purpose metric for comparing theories. A simple formal criterion for distinguishing fair dilemmas from unfair ones would allow us to more systematically test the claim that FDT outperforms the competition on all fair dilemmas.

We do not yet know whether FDT is optimal among decision theories, or even (on a formal level) what optimality consists in. Our initial results, however, suggest that FDT is an important advance over CDT and EDT. If there are better decision theories, we expect them to extend and enrich FDT's approach, rather than returning to CDT or EDT's characteristic methods for constructing hypotheticals and selecting actions.

\todo[color=yellow,inline]{Acknowledgements: We acknowledge Rob Bensinger and Ben Levenstein for extensive help with the presentation of this \paper, and for many valuable conversations. Functional decision theory has been developed in many parts by the co-authors and other contributors in a discussion ongoing since 2010 and earlier; we acknowledge Wei Dai, Vladimir Nesov, Vladimir Slepnev, and Patrick LaVictoire in particular for their contributions.}

\printbibliography
\end{document}